\documentclass[10pt,twocolumn,letterpaper]{article}

\usepackage[pagenumbers]{cvpr} 

\definecolor{cvprblue}{rgb}{0.21,0.49,0.74}
\usepackage[pagebackref,breaklinks,colorlinks,allcolors=cvprblue]{hyperref}
\usepackage{xcolor}         
\usepackage[accsupp]{axessibility} 

\usepackage{enumitem}  
\usepackage{graphicx}
\usepackage{multirow}
\usepackage{colortbl}
\usepackage[table]{xcolor}

\usepackage{bbm}

\usepackage{float}

\def\paperID{9873} 
\def\confName{CVPR}
\def\confYear{2026}

\title{PosterIQ: A Design Perspective Benchmark for \\Poster Understanding and Generation}

\author{
    Yuheng Feng\textsuperscript{1} \quad
    Wen Zhang\textsuperscript{2} \quad
    Haodong Duan\textsuperscript{3} \quad
    Xingxing Zou\textsuperscript{1 $^*$} \vspace{0.5em} \\
    \textsuperscript{1}The Hong Kong Polytechnic University \quad 
    \textsuperscript{2}Snapchat Inc. \quad  
    \textsuperscript{3}ByteDance Seed \quad
    \textsuperscript{$^*$}Corresponding author \\
    {\tt\small bruce.feng@connect.polyu.hk, wenzhang.ccm@gmail.com,} \\
    \vspace{-1mm}
    {\tt\small dhd.efz@gmail.com, xingxing.zou@polyu.edu.hk}\\
}

\begin{document}
\maketitle
\begin{abstract}
We present PosterIQ, a design-driven benchmark for poster understanding and generation, annotated across composition structure, typographic hierarchy, and semantic intent. It includes 7,765 image–annotation instances and 822 generation prompts spanning real, professional, and synthetic cases.
To bridge visual design cognition and generative modeling, we define tasks for layout parsing, text–image correspondence, typography/readability and font perception, design quality assessment, and controllable, composition-aware generation with metaphor. We evaluate state-of-the-art MLLMs and diffusion-based generators, finding persistent gaps in visual hierarchy, typographic semantics, saliency control, and intention communication; commercial models lead on high-level reasoning but act as insensitive automatic raters, while generators render text well yet struggle with composition-aware synthesis.
Extensive analyses show PosterIQ is both a quantitative benchmark and a diagnostic tool for design reasoning, offering reproducible, task-specific metrics. We aim to catalyze models' creativity and integrate human-centred design principles into generative vision–language systems. \url{https://github.com/ArtmeScienceLab/PosterIQ-Benchmark}
\end{abstract}
\section{Introduction}

Multimodal large language models (MLLMs) \cite{comanici2025gemini, openai_gpt5_2025,anthropic_claude_sonnet_45_2025} have recently advanced in visual understanding—from object and scene recognition to cross-modal alignment, fine-grained parsing, and open-vocabulary detection—enabling high-level semantics and robust reasoning in complex contexts. In parallel, generative models have progressed across text-to-image, image-to-image, and interactive co-creation, with emerging strengths in style control, layout composition, and semantic consistency. These gains are especially evident in creative applications, where models show tangible innovation and practical utility in advertising, branding, and visual metaphor, while LLMs increasingly aid story ideation, tone transfer and narrative structuring.

Despite benchmarks like ~\cite{fang2025creation}, evaluations remain largely text-centric. Image-generation assessments often prioritize aesthetics and overlook the compositional, constraint-driven nature of design. This omission is most acute in posters—a tightly integrated medium where visual understanding and content generation must align under strict constraints. Here, theme interpretation, information hierarchy, typographic rules, text–image coupling, theme consistency, and audience preference interact in ways single-dimensional metrics fail to capture as “genuine creativity.” Posters are not merely about visual appeal; they are visual communication media designed to ensure key messages are perceived, understood, and remembered with minimal cognitive load. Design without communicative purpose yields only superficial elegance. Effective poster design integrates robust text recognition, semantic understanding, faithful rendering, and hierarchical layout to keep critical elements legible in dense compositions; typographic choices must balance readability and beauty; overall style must align with audience and theme; and visual devices such as metaphor, symbolism, and whitespace should reinforce messages and aid memory through visual rhetoric. Accordingly, key assessment dimensions include accurate text understanding, goal-directed typography and layout, coherent text–image coordination, style control under audience and thematic constraints, and creative expression through metaphor—all grounded in understanding of design rules and theory, plus good taste and creative thinking, whose integration separates strong design from the merely adequate.

From a modeling perspective, poster-oriented understanding and generation systems must jointly satisfy: (1) text understanding and readability via robust OCR and font recognition, readability prediction (glyph shape, weight, size, contrast) with constraint-aware optimization, and automatic text hierarchy and structured layout; (2) layout reasoning and hierarchical organization through explicit modeling of grids, whitespace, alignment, layering, and figure–ground relationships for global optimization across elements, sizes, and densities; (3) semantic–style consistency with task- and audience-aware style retrieval or transfer under target constraints, suppressing style noise that could obscure key information; (4) text–image coordination and saliency control that transforms key messages into semantically aligned visuals and directs attention so core information is perceived first; and (5) rhetorical modeling and metaphor generation that produce decodable visual metaphors via semantic association and analogy while balancing novelty against misinterpretation risks. 

\begin{figure*}
    \centering
    \includegraphics[width=1\linewidth]{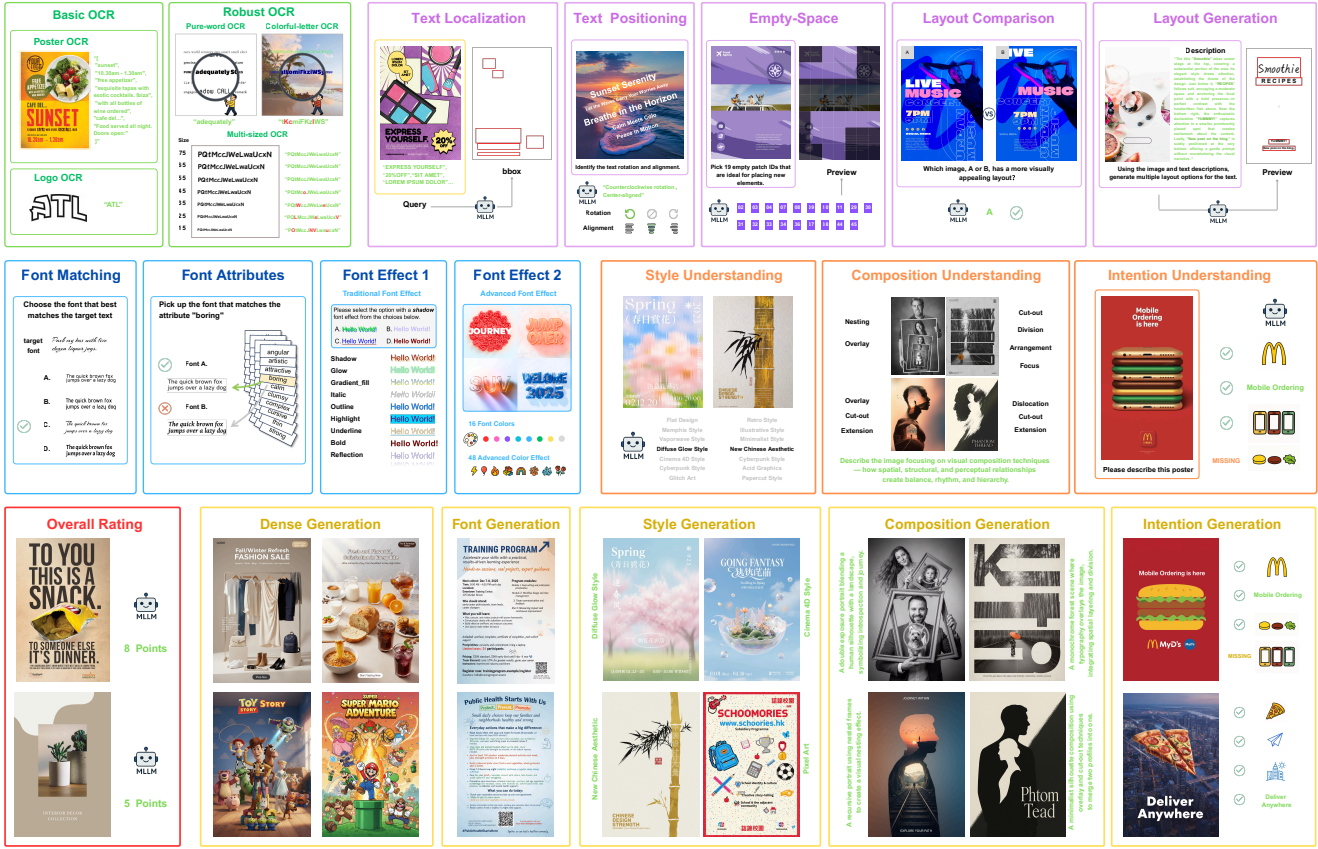}
    \caption{Overview of the benchmark, which includes over a dozen tasks}
    \label{fig:benchmark}
    \vspace{-5mm}
\end{figure*}

To address this, we introduce PosterIQ: a systematic benchmark for highly constrained poster creation that spans the full pipeline—from understanding and ideation to composition and generation—and more comprehensively characterizes MLLMs’ creative capabilities in poster understanding and synthesis. As shown in Fig.~\ref{fig:benchmark}, it comprises two tightly coupled, design–oriented components: an understanding module with a global quality assessment (overall rating) and four decoupled task families—(i) OCR and text readability, (ii) font shape perception and attribute understanding, (iii) multidimensional layout and hierarchy perception, and (iv) high-level style recognition, visual deconstruction, and semantic communication—and a generation module that exceeds existing benchmarks by (i) generating and organizing high-density visual content, (ii) accurately generating dense text with diverse fonts, (iii) enabling controllable poster style and thematic tone, (iv) supporting challenging structural decomposition and recomposition of visual elements, and (v) facilitating intention communication and creative evaluation via metaphor and visual rhetoric; overall, PosterIQ advances verifiable creativity in real-world design scenarios through actionable tasks, reproducible metrics, and decoupled evaluation dimensions. From a broad empirical analysis, it reveals systematic gaps between open-source and proprietary models. Frontier commercial systems generally perform better on high-level tasks such as layout reasoning, composition understanding, and intention interpretation. However, when used as automatic raters for poster quality, their scores often lack sensitivity and fail to clearly separate strong and weak designs. In the generation setting, current models already demonstrate strong visual synthesis and text rendering capabilities, but still struggle with composition-aware generation and clear expression of design intentions. A cross-model comparison further shows noticeable differences in the richness and diversity of generated typography across systems.
All in all, our contributions are:
\begin{itemize}
    \item \textbf{Data.} We build a comprehensive poster-centric benchmark with 7{,}765 image–annotation instances for understanding and 822 prompts for generation, combining real-world posters, professionally designed layouts, and synthetic cases for typography, layout, and metaphor.
    \item \textbf{Task Coverage.} We provide an in-depth and systematic evaluation of poster-related abilities, covering OCR robustness, font perception, multi-dimensional layout and spatial reasoning, high-level style and intention understanding, as well as dense, style-aware, composition-aware, and metaphor-driven generation.
    \item \textbf{Evaluation Benchmark.} We introduce PosterIQ as a rigorous, task-specific evaluation framework with reproducible metrics tailored to each challenge. Our analysis offers fine-grained insights into accuracy, robustness, aesthetic judgement, and creative intent.
\end{itemize}

\section{Related Work}

\subsection{Multimodal Benchmarks}
A growing range of benchmarks evaluates MLLMs across diverse tasks~\cite{hendrycks2021measuring, rein2024gpqa, wang2024mmlu,chen2021evaluating,jain2024livecodebench}.  Concurrently, some studies explicitly investigate cross-modal diversity and complex semantic alignment~\cite{wen2023dip,wen2025domain,hu2026ai}. As visual modality emerged, vision–language benchmarks such as MMBench~\cite{liu2024mmbench}, Creation-MMBench~\cite{fang2025creation}, and Seed-Bench-2-Plus~\cite{li2024seed} began testing multimodal understanding and instruction following from everyday recognition to complex reasoning. OCR-focused suites~\cite{liu2024ocrbench,fu2024ocrbench} target scene text recognition and structured extraction. Image generation is now systematically assessed: ELLA’s DPG measures fine-grained prompt adherence under dense descriptions~\cite{hu2024ella}, while GenEval evaluates text–image alignment for multi-object, multi-attribute, and relational prompts~\cite{ghosh2023geneval}. For text rendering within images, TextCrafter’s CVTG-2K tests readability and fidelity in complex scenes~\cite{du2025textcrafter}, and X-Omni’s LongText-Bench (English/Chinese) probes paragraph-level rendering and layout consistency~\cite{geng2025x}. Specialized evaluations cover narrower domains—VGBench~\cite{zou2024vgbench}, ChartQA~\cite{masry2022chartqa}, and InfographicVQA~\cite{mathew2022infographicvqa} for vector graphics and chart reasoning; LiveIdeaBench~\cite{ruan2024liveideabench} and DesignProbe~\cite{lin2024designprobe} for ideation-centric evaluation; and UI-Vision~\cite{nayak2025ui} for desktop GUIs. 

In contrast, PosterIQ adopts a visual design perspective, emphasizing creative poster design, making it more challenging and closer to real-world design practice.

\subsection{Poster Design and Generative Models}
Diffusion models are emerging as the leading approach for high-quality image synthesis~\cite{zhao2025lex, stablediffusion3, flux2024}, lowering the barrier for non-experts to engage in creative design. Beyond static images, diffusion-based methods support controllable pipelines for structured manipulation of layout, typography, and style. Recent multimodal image generators—Qwen-Image~\cite{wu2025qwen}, Seedream 4.0~\cite{seedream4}, GPT-Image-1~\cite{openai_gpt_image_1}, and Gemini 2.5 Flash Image~\cite{gemini_2_5_flash_image}—improve efficiency in T2I and image-editing. As these models advance, interest grows in  poster design. Hierarchical frameworks like instruction-tuned models like PosterLLaVA~\cite{yang2024posterllava} and COLE~\cite{jia2023cole} enable automatic poster generation, while VASCAR~\cite{zhang2024vascar} targets content-aware layout. FontCLIP~\cite{tatsukawa2024fontclip}, ControlText~\cite{jiang2025controltext}, and FontTS~\cite{shi2024fonts} advance semantic typography and font control, building on early crowdsourced studies of font attributes~\cite{o2014exploratory}. Pipeline-based automation is realized in systems such as OpenCOLE~\cite{inoue2024opencole}. PosterCraft~\cite{chen2025postercraft} uses Gemini to generate training captions, and DreamPoster~\cite{hu2025dreamposter} uses a specialized captioner for glyph- and layout-level descriptions.
Our benchmark shifts the focus to the evaluation of understanding and generation in poster design.

\section{Benchmark}

\subsection{Understanding Tasks}
\textbf{OCR Tasks.}
To support poster understanding and design-oriented reasoning, MLLMs must first show reliable visual text recognition. Our benchmark targets poster-specific traits and evaluates OCR across five sub-tasks: \textit{1) Logo OCR—recognizing} highly stylized, distorted, or abstracted logo typography with irregular character shapes; \textit{2) Real-World Poster OCR}—handling complex scenes with diverse fonts, scales, dense layouts, and textured backgrounds where text–graphic interactions raise difficulty; \textit{3) Simple OCR}—estimating upper-bound performance by rendering Oxford 3000 words with varied fonts and casing as length-balanced sequences on white backgrounds; \textit{4) Hard OCR}—testing robustness via unordered letter sequences (excluding ambiguous “l”/“I”) rendered in varied fonts, slight rotations, random colors, and placed on highly textured, colorful backgrounds from real images; \textit{5) Multi-Size OCR}—assessing stability under scale variation by generating unordered letter sequences in 14 font sizes with a base font on white, repeated across multiple fonts.

\textbf{Font Understanding Tasks.}

Typography is central to poster design: font styles convey both theme and aesthetics, and require accurate perception. We evaluate these capabilities with four sub-tasks: \textit{1) Font Matching}—fine-grained style identification without font-name priors; given a target text in one font, the model selects the matching font from nine candidates showing different text, forcing reliance on visual style rather than character identity; \textit{2) Font Attribute Perception}—using 37 human-derived attributes from ~\cite{o2014exploratory}. (31 relative, 6 binary such as serif/italic), the model sees a pair of font-rendered texts and chooses the one best matching a target attribute, with accuracy reflecting agreement with human judgments; \textit{3) Traditional Font Effect Recognition}—recognition of nine common effects (bold, italic, underline, etc.) by selecting the correct effect from four options; \textit{4) Advanced Font Effect Recognition}—on highly stylized, model-generated and curated effects, the model first identifies the text’s dominant color, then selects the correct effect from 48 candidates spanning styles such as rock.

\textbf{Spatial Reasoning Tasks.}
Effective perception and generation of layout structures are vital for poster design, reflecting an MLLM’s grasp of aesthetics, spatial balance, and composition. We evaluate layout understanding, aesthetic judgment, and spatial planning with five sub-tasks: \textit{1) Text Localization}—given a list of target phrases, the model returns a bounding box for each; coordinates are normalized to [0,1] to handle varying resolutions and aspect ratios, emphasizing precise detection of small or dense text; \textit{2) Text Alignment and Rotation}—the model infers alignment and rotation for each text sample, capturing typographic structure beyond position; \textit{3) Empty-Space Perception}—using partially completed posters divided into a $7\times7$ grid, annotators label suitable regions for new elements; their intersection defines the consensus empty space, and the model outputs region IDs for a requested count, evaluated by IoU with the consensus set; \textit{4) Layout Comparison}—given a professional layout and a version deliberately violating design principles, the model selects the more coherent and aesthetically sound design, testing high-level layout judgment; \textit{5) Layout Generation}—given a textual specification of relative sizes and placements ($\leq 5$ elements), the model produces bounding boxes forming a coherent layout, evaluated on positional and area accuracy, thereby assessing instruction following and multimodal reasoning.

\textbf{Advanced Visual Design Understanding.}
High-quality poster design goes beyond basic layout and elements, often employing sophisticated styles, visual deconstruction, and conceptual metaphors. We evaluate advanced design understanding with three sub-tasks: \textit{1) Poster Style Classification}—identify one of 17 curated styles (e.g., Minimalist, Diffuse Glow, Memphis) based on holistic cues spanning color, effects, typography, and imagery; \textit{2) Composition Structure Understanding}—describe each poster’s visual construction using operations such as misalignment, segmentation, nesting, cutouts, repetition, extension, focus shifting, and mirroring, with coverage judged by an LLM against human-annotated key concepts; \textit{3) Intention and Metaphor Interpretation}—explain metaphor-rich posters (e.g., stacked smartphones evoking a hamburger, toy soldiers forming a dove, circuit-board textures fused with leaf veins), with success measured by capturing manually annotated explanatory elements and intended messages.

\textbf{Rating Task.}
Finally, we further assess aesthetic judgment by asking models to assign each poster a quality score from 0 to 10. After normalizing scores, we measure alignment with human preferences by computing the correlation between model predictions and human rating distributions.

\subsection{Generation Tasks}

\textbf{Dense Generation.}
Real-world themes (e.g., film IP) demand many characters, objects, and fine details in one image. We curate themes with ~10 entities and specify actions, attributes, or orientations for each; given a prompt, the model must render all required elements. An MLLM verifies each item and reports overall matching accuracy.

\textbf{Font Generation.}
 To assess control over typography, we ask for text-centric posters across varied scenarios while explicitly encouraging font diversity. An MLLM infers latent font attributes (e.g., bold, elegant, friendly), and the count of distinct attributes measures font-style diversity.

\textbf{Style Generation.}
The model generates posters in 17 mainstream styles (matching our style-understanding benchmark, e.g., Memphis). An MLLM predicts each poster’s style, and accuracy is computed against the target.

\textbf{Composition Generation.}
Advanced design often employs structured compositional techniques—such as misalignment, segmentation, nesting, cutouts, repetition, extension, focus shifting, and mirroring—to create visual tension and sophistication. 
We provide prompts describing these composition strategies and require the model to generate images accordingly. 
An MLLM evaluates whether each compositional element appears in the output, using the same criteria as in the composition-understanding task, and computes matching rate across all required points.

\textbf{Intention Generation.}
We prompt metaphor- and concept-rich posters (e.g., dual-meaning imagery or abstract symbols). An MLLM evaluates whether key intention elements—essential visual cues and intended semantics, as defined by human annotations—are present, reusing the labels from the understanding task.

\section{Experiment}
\newcolumntype{C}[1]{>{\centering\arraybackslash}p{#1}}

\begin{table*}[t]
\centering
\caption{Comprehensive OCR benchmark results across multiple visual text recognition tasks.
$AC$ denotes the accuracy, and $WR$ denotes the word-level recall rate. $\Delta$ represents the performance gap between the simple and hard OCR settings, reflecting model robustness. $Std$ denotes the standard deviation of $WR$ across different font sizes, indicating stability.}
\vspace{-2mm}
\renewcommand{\arraystretch}{1.1}

\resizebox{1\textwidth}{!}{%
\begin{tabular}{|C{1.2cm}|p{3cm}|C{2.8cm}|C{2.8cm}|C{2.8cm}|C{2.8cm}|C{2.3cm}|C{2.2cm}|C{2.2cm}|}
\toprule
\rowcolor{gray!10}
\multicolumn{2}{|c|}{} & \multicolumn{1}{c|}{\textbf{Logo OCR}} & \multicolumn{1}{c|}{\textbf{Poster OCR}} & \multicolumn{1}{c|}{\textbf{Simple OCR}} & \multicolumn{1}{c|}{\textbf{hard OCR}} & \multicolumn{1}{c|}{$\Delta$}  & \multicolumn{2}{c|}{\textbf{Font Size OCR}}\\
\cline{3-9}
\rowcolor{gray!10}
\multicolumn{2}{|c|}{\multirow{-2}{*}{\cellcolor{gray!10}\textbf{Model}}} & $AC$ $\uparrow$ & $AC$ $\uparrow$ & $WR$ $\uparrow$  & $WR$ $\uparrow$  & $\delta$ $\downarrow$ & $WR$ $\uparrow$ & $Std$ $\downarrow$ \\
\hline

\multirow{4}{*}{\textbf{Closed}} 
& GPT-5 & \textbf{0.952} & 0.922 & 0.965 & 0.496 & 0.469 & \textbf{0.885} & 0.113 \\
& Claude-Sonnet-4.5 & 0.902 & 0.884 & 0.951 & 0.579 & 0.372 & 0.878 & 0.035 \\
& Gemini-2.5-Pro & 0.923 & \textbf{0.952} &\textbf{0.997} & 0.472 & 0.525 & 0.879 & 0.037 \\
& Grok-4-fast & 0.440 & 0.834 & 0.769 & 0.044 & 0.725 & 0.288 & 0.105 \\

\hline
\multirow{4}{*}{\textbf{Open}}

& MiniCPM-V-4.5 & 0.883 & 0.932 & 0.989 & 0.521 & 0.468 & 0.865 & \textbf{0.023} \\
& Gemma-3n-e4b-it & 0.895 & 0.891 & 0.991 & 0.231 & 0.760 & 0.712 & 0.115 \\
& Qwen3-VL-4B & 0.887 & 0.921 & 0.963 & 0.726 & 0.237 & 0.679 & 0.196 \\
& Qwen3-VL-8B & 0.882 & 0.931 & 0.937 & \textbf{0.781} & \textbf{0.156} & 0.676 & 0.242 \\

\bottomrule
\end{tabular}%
}
\vspace{-5mm}
\label{tab:ocr_results}
\end{table*}

\textbf{Data.} PosterIQ consists of 7{,}765 annotated instances for understanding and 822 prompts for generation, spanning 24 task types in total. The understanding part covers five OCR tasks (3{,}005 items: logo, poster, simple, hard, and font-size OCR), four font perception tasks (2{,}788 items: font matching, font attributes, and two levels of font effects), six spatial reasoning tasks (1{,}178 items: text localization, rotation, alignment, empty-space perception, layout comparison, and layout generation), three advanced visual design tasks (575 items: style, composition, and intention understanding), and an overall rating task (219 items). The generation part includes five task families with 822 prompts: dense content generation (114), font generation (135), style generation (256), composition generation (117), and intention generation (200), jointly supporting a comprehensive evaluation of both poster understanding and generation.

\textbf{Metric.} We use task-specific metrics to capture both accuracy and robustness.
For \textit{Logo OCR} and \textit{Poster OCR}, we report item-level accuracy. For \textit{Simple}, \textit{Hard}, and \textit{Font-Size OCR}, we use a word-level recall rate (\emph{WR}) over fixed-length segments, together with the standard deviation (\emph{Std}) across font sizes and the gap $\Delta$ between simple and hard settings.
For the \textit{font tasks}, we adopt a normalized multiple-choice score (\emph{Score}) in $[0,1]$: values near 0 indicate near-random predictions, and 1 indicates all answers are correct.
For \textit{layout tasks}, we use IoU-based metrics and discrete-choice accuracy. \emph{Top-1 IoU} denotes the maximum IoU over predicted regions, while \textit{Alignment} and \textit{Rotation} are evaluated by multiple-choice accuracy. In the \textit{Empty Space} task, \emph{Matching Acc.} measures whether the returned region IDs match the requested count and target regions. For \textit{Layout Generation}, we measure the center offset and area ratio between predicted and ground-truth boxes.
For \textit{Advanced Visual Design Understanding}, \emph{Score} is again a multiple-choice accuracy, and \emph{Points Score} measures the coverage of annotated key points in the model’s description. In the \textit{overall rating} task, we compute the cosine similarity between the vector of model scores and the vector of human ratings.
Generation experiments are evaluated with analogous metrics (e.g., option correctness, style agreement, and key-point coverage); find full metric definitions in the supplementary material.

\subsection{Understanding Task}

\textbf{OCR Tasks.}
As shown in Tab.~\ref{tab:ocr_results}, all models, except Grok-4-fast, achieve accuracy close to 0.9 on both Logo OCR and Poster OCR, indicating that most MLLMs can reliably handle standard text recognition in design-oriented imagery. 
In the synthetic setting, the gap between the simple and hard OCR reveals notable differences in robustness: Claude-Sonnet-4.5 and the Qwen family exhibit the most stable performance under heavy visual interference. 
Meanwhile, Claude-Sonnet-4.5\cite{anthropic_claude_sonnet_45_2025}, Gemini-2.5-Pro\cite{comanici2025gemini}, and MiniCPM-V-4.5\cite{yu2025minicpm} show minimal performance degradation across scales, suggesting stronger invariance to font-size variation.

\begin{table*}[t]
    \centering
    \begin{minipage}[t]{0.505\textwidth}
        \centering
\renewcommand{\arraystretch}{1.2}
\centering
\caption{Comparison of Font tasks across models.}
\vspace{-2mm}

\resizebox{1\textwidth}{!}{%
\small

\begin{tabular}{|C{1cm}|p{2.5cm}|C{2.1cm}|C{2.1cm}|C{2.1cm}|C{1.8cm}|C{1.8cm}|}
\toprule
\rowcolor{gray!10}
\multicolumn{2}{|c|}{} & \multicolumn{1}{c|}{\textbf{Font Matching}} & \multicolumn{1}{c|}{\textbf{Font Attributes}} & \multicolumn{1}{c|}{\textbf{Font Effect 1}} & \multicolumn{2}{c|}{\textbf{Font Effect 2}} \\
\cline{3-7}
\rowcolor{gray!10}
\multicolumn{2}{|c|}{\multirow{-2}{*}{\cellcolor{gray!10}\textbf{Model}}}
& Score $\uparrow$ & Score $\uparrow$ & Effect Score $\uparrow$ & Color Score $\uparrow$ & Effect Score $\uparrow$ \\
\hline

\multirow{4}{*}{\cellcolor{white!100}\textbf{Closed}}
& GPT-5              & 0.668 & 0.664 & \textbf{0.805} & 0.753 & 0.189 \\
& Claude-Sonnet-4.5  & \textbf{0.699} & 0.633 & 0.773 & 0.710 & 0.204 \\
& Gemini-2.5-Pro     & 0.362 & \textbf{0.720} & 0.790 & \textbf{0.804} & \textbf{0.358} \\
& Grok-4-fast        & 0.044 & 0.559 & 0.525 & 0.727 & 0.247 \\
\hline

\multirow{4}{*}{\textbf{Open}}
& MiniCPM-V-4.5          & -0.001 & 0.653 & 0.555 & 0.718 & 0.181 \\
& Gemma-3n-e4b-it        & -0.012 & 0.603 & 0.395 & 0.701 & 0.218 \\
& Qwen3-VL-4B   & 0.083  & 0.645 & 0.575 & 0.770 & 0.330 \\
& Qwen3-VL-8B   & 0.063  & 0.607 & 0.565 & 0.761 & 0.229 \\

\bottomrule
\end{tabular}%
\label{Tab:font tasks}
}
    \end{minipage}
    \hfill
    \begin{minipage}[t]{0.485\textwidth}
        \centering
\renewcommand{\arraystretch}{1.2}
\centering
\caption{Results of Understanding.}
\vspace{-2mm}

\resizebox{1\textwidth}{!}{%
\small
\begin{tabular}{|C{1cm}|p{2.5cm}|C{2.8cm}|C{3.7cm}|C{3.4cm}|}
\toprule
\rowcolor{gray!10}
\multicolumn{2}{|c|}{} & \multicolumn{1}{c|}{\textbf{Style Understanding}} & \multicolumn{1}{c|}{\textbf{Composition Understanding}} & \multicolumn{1}{c|}{\textbf{Intention Understanding}}\\
\cline{3-5}
\rowcolor{gray!10}
\multicolumn{2}{|c|}{\multirow{-2}{*}{\cellcolor{gray!10}\textbf{Model}}} 
                                    & Score $\uparrow$ & Points Score $\uparrow$ & Points Score $\uparrow$\\
\hline

\multirow{4}{*}{\cellcolor{white!100}\textbf{Closed}}
& GPT-5  & \textbf{0.851} & 0.730 & \textbf{0.824} \\
& Claude-Sonnet-4.5 & 0.813 & 0.608 & 0.761 \\
& Gemini-2.5-Pro & 0.830 & \textbf{0.802} & 0.788 \\
& Grok-4-fast & 0.560 & 0.717 & 0.771 \\
\hline

\multirow{4}{*}{\textbf{Open}}

& MiniCPM-V 4.5 & 0.631 & 0.635 & 0.691 \\
& Gemma-3n-e4b-it & 0.514 & 0.504 & 0.598 \\
& Qwen3-VL-4B & 0.805 & 0.672 & 0.701 \\
& Qwen3-VL-8B & 0.610 & 0.684 & 0.710 \\

\bottomrule
\end{tabular}%
\label{Tab:ads}
}
\end{minipage}
\vspace{-2mm}
\end{table*}

\textbf{Font Understanding Benchmark Analysis.}
Across the four font-related tasks, we observe large capability gaps among current MLLMs from Tab.~\ref{Tab:font tasks}. 
In the font matching task, only GPT-5, Claude-Sonnet-4.5, and Gemini-2.5-Pro demonstrate meaningful discrimination of typographic styles, while most other models perform at a near-chance level. 
For perceptual font attributes, Gemini-2.5-Pro aligns most closely with human judgments, followed by GPT-5. 
In recognizing traditional and advanced font effects—including style-specific color cues—Gemini-2.5-Pro consistently achieves the strongest overall performance.

\textbf{Advanced Visual Design Understanding.}
On the style understanding task in Tab.~\ref{Tab:ads}, GPT-5, Claude-Sonnet-4.5, Gemini-2.5-Pro, and Qwen3-VL-4B correctly identify over 80\% of poster styles, indicating that current MLLMs can reliably capture global stylistic cues. 
For more complex tasks, Gemini-2.5-Pro achieves the best performance on visual composition understanding (0.802), while GPT-5 performs best on intention understanding (0.824). 
These suggest that proprietary models currently exhibit stronger design literacy than open-source counterparts, especially when higher-level composition and conceptual intent are involved.

\begin{table*}[t]
    \centering
        \centering  
\renewcommand{\arraystretch}{1.2}
\centering

\caption{Comparison of layout reasoning tasks across models.}
\vspace{-3mm}
\resizebox{1\textwidth}{!}{%
\small
\begin{tabular}{|C{1cm}|p{3cm}|c|c|c|c|c|c|c|c|c|c|c|c|}
\toprule
\rowcolor{gray!10}
\multicolumn{2}{|c|}{} & \multicolumn{4}{c|}{\textbf{Text Localization}}& \multicolumn{2}{c|}{\textbf{Text Positioning}}& \multicolumn{2}{c|}{\textbf{Empty-Space}}& \multicolumn{1}{c|}{\textbf{Layout Comparison}}& \multicolumn{3}{c|}{\textbf{Layout Generation}}\\
\cline{3-14}
\rowcolor{gray!10}
\multicolumn{2}{|c|}{\multirow{-2}{*}{\cellcolor{gray!10}\textbf{Model}}} 
& Top-1 IoU $\uparrow$ & Top-3 IoU $\uparrow$ & Mean IoU $\uparrow$ & Recall $\uparrow$ 
& Alignment $\uparrow$ & Rotation $\uparrow$ 
& Mean IoU $\uparrow$ & Match Acc. $\uparrow$ 
& Score $\uparrow$ 
& Center Bias $\downarrow$ & Area Ratio $\uparrow$ & Recall $\uparrow$ \\
\hline

\multirow{4}{*}{\textbf{Closed}}
& GPT-5               & 0.432 & 0.308 & 0.171 & 0.971 & 0.347 & 0.480 & 0.384 & 0.820 & \textbf{0.719} & \textbf{0.084} & 0.484 & 1.000 \\
& Claude-Sonnet-4.5   & 0.163 & 0.104 & 0.060 & 0.905 & \textbf{0.468} & 0.488 & 0.300 & 0.976 & 0.648 & 0.145 & 0.423 & 0.953 \\
& Gemini-2.5-Pro      & \textbf{0.491} & 0.404 & 0.295 & 0.899 & 0.325 & \textbf{0.576} & \textbf{0.491} & \textbf{0.982} & 0.680 & \textbf{0.084} & \textbf{0.569} & 1.000 \\
& Grok-4-fast         & 0.087 & 0.065 & 0.033 & 0.978 & 0.205 & 0.129 & 0.241 & 0.784 & 0.172 & 0.130 & 0.439 & 1.000 \\
\hline

\multirow{4}{*}{\textbf{Open}}
& MiniCPM-V-4.5         & 0.269 & 0.228 & 0.150 & \textbf{0.992} & 0.017 & 0.232 & 0.280 & 0.407 & 0.305 & 0.243 & 0.376 & 0.991 \\
& Gemma-3n-e4b-it       & 0.182 & 0.115 & 0.056 & 0.974 & -0.087 & 0.254 & 0.249 & 0.689 & 0.086 & 0.259 & 0.352 & 1.000 \\
& Qwen3-VL-4B  & 0.430 & 0.368 & 0.223 & 0.819 & 0.130 & 0.290 & 0.264 & 0.778 & 0.633 & 0.250 & 0.337 & 0.955 \\
& Qwen3-VL-8B  & 0.741 & \textbf{0.680} & \textbf{0.450} & 0.978 & 0.152 & 0.495 & 0.230 & 0.874 & 0.266 & 0.214 & 0.394 & 0.917 \\
\bottomrule
\end{tabular}%
}
\label{Tab:layout_task}
\vspace{-5mm}
\end{table*}

\begin{figure*}[t]
    \centering
    \includegraphics[width=1\textwidth]{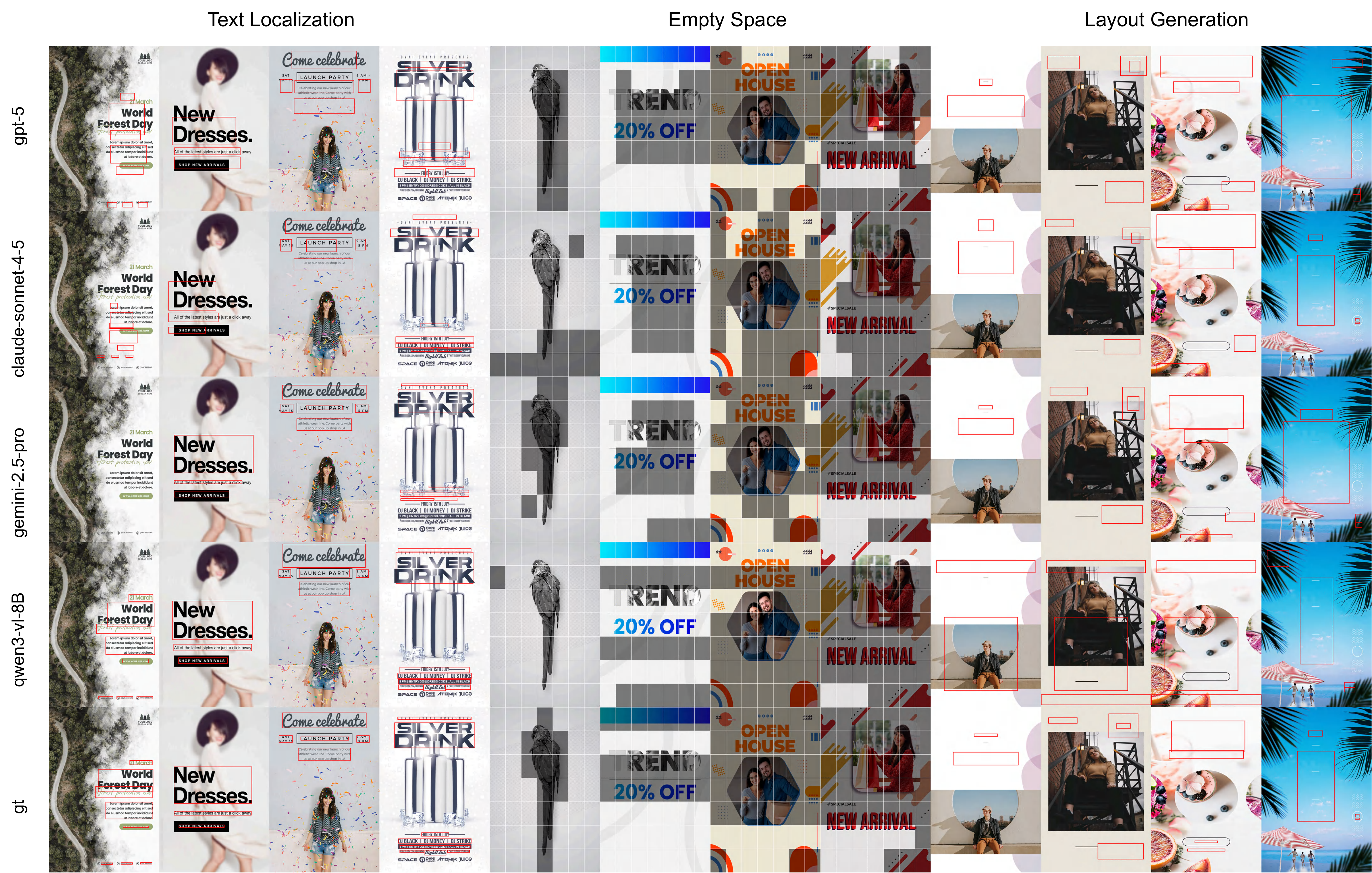}
    \vspace{-7mm}
    \caption{Qualitative comparison of four models on three layout-related tasks. 
For \textit{Text Localization} and \textit{Layout Generation}, the predicted bounding boxes are shown in red. 
For the \textit{Empty Space} task, the selected patch IDs are highlighted in the image.}
    \vspace{-5mm}
    \label{fig:grounding}
\end{figure*}

\textbf{Layout Tasks.}
Our layout benchmark primarily tests spatial recognition and layout intuition (Tab.~\ref{Tab:layout_task}).
On the \textit{Text Localization}, Qwen3-VL-8B achieves the best performance among open-source models, with a mean IoU of 0.45, followed by the Closed Gemini-2.5-Pro (0.295), as also illustrated in Fig.~\ref{fig:grounding}.
For \textit{Text Position} (alignment), Closed models generally outperform open-source ones, and Gemini-2.5-Pro further leads on \textit{Text Rotation} recognition.
In the \textit{Empty Space} task, which requires spatial reasoning over potential placement regions, Gemini-2.5-Pro again shows superior performance.
For \textit{Layout Comparison}, GPT-5, Claude-Sonnet-4.5, and Gemini-2.5-Pro consistently outperform other models.
Finally, \textit{Layout Generation} requires joint reasoning over visual context and textual instructions; here, Closed models overall outperform open-source counterparts, with Gemini-2.5-Pro achieving best results.

\begin{table}[t]
    \centering
        \centering  
\renewcommand{\arraystretch}{1.2}
\centering

\caption{Comparison of Overall Rating}
\vspace{-2mm}
\resizebox{0.48\textwidth}{!}{%
\small
\begin{tabular}{|C{3cm}|p{5cm}|C{5cm}|}
\toprule
\rowcolor{gray!10}
\multicolumn{2}{|c|}{} & \multicolumn{1}{c|}{\textbf{Overall Rating }} \\
\cline{3-3}
\rowcolor{gray!10}
\multicolumn{2}{|c|}{\multirow{-2}{*}{\cellcolor{gray!10}\textbf{Model}}} 
                                    & Sim $\uparrow$ \\
\hline

\multirow{4}{*}{\cellcolor{white!100}\textbf{Closed}}
& GPT-5 & 0.347 \\
& Claude-Sonnet-4.5 & 0.384 \\
& Gemini-2.5-Pro & 0.399 \\
& Grok-4-fast & 0.465 \\
\hline

\multirow{4}{*}{\textbf{Open}}
& MiniCPM-V 4.5 & 0.095 \\
& Gemma 3N-E4B-IT & \textbf{0.483} \\
& Qwen3-VL-4B & 0.172 \\
& Qwen3-VL-8B & 0.237 \\
\bottomrule
\end{tabular}%
}
\label{Tab:rating_task}
\vspace{-6mm}
\end{table}

\textbf{Overall Rating.}
Asking MLLMs to assign a single holistic quality score to posters is a challenging task. 
Although most models can produce reasonable scores, Tab.~\ref{Tab:rating_task} shows that their predicted ratings exhibit relatively low correlation with human judgments. 
This gap highlights the importance of our decoupled evaluation setting, where fine-grained understanding and design dimensions are assessed separately rather than relying solely on a global score.

\vspace{-1mm}
\subsection{Generation Task}
\vspace{-1mm}
We evaluate four models (Seedream-4.0\cite{seedream4}, Gemini-2.5-Flash-Image\cite{gemini_2_5_flash_image}, GPT-Image-1\cite{openai_gpt_image_1}, Qwen-Image\cite{wu2025qwen}) on five generation tasks: Dense Generation, Font Generation, Style Generation, Composition Generation, and Intention Generation.

\textbf{Quantitative Results.}
Averaged across five tasks, Gemini-2.5-Flash-Image leads overall. Per-task results show differentiated strengths: Gemini excels in Composition and Intention, indicating strong global layout planning and instruction–semantics alignment, while GPT-Image-1 leads in Style and Intention but trails on Dense and Font, revealing weaknesses in micro-structure fidelity and text readability. Structural patterns emerge: Font and Dense are partially decoupled (e.g., Seedream-4.0 is strong on Dense but middling on Font), implying distinct capability axes and a need for targeted supervision on stroke closure, spacing, ligatures, and variant-shape modeling. A global–local trade-off is evident—models dominating Composition and Intention generally do not lead on Style and Font—suggesting training favors global planning over local precision. Font remains the bottleneck, with uniformly low scores (best 0.391) versus high Composition (up to 0.866), posing risks for multilingual scripts, small sizes, complex fonts, and low-contrast backgrounds.

\begin{table}[t]
\centering
\caption{Performance of image generation models across five evaluation tasks. \textbf{Point Score}($PS$), \textbf{Score}($S$), \textbf{Richness}($R$)}
\vspace{-2mm}
\renewcommand{\arraystretch}{1.2}
\resizebox{0.48\textwidth}{!}{%
\begin{tabular}{|p{4.5cm}|c|c|c|c|c|c|}
\toprule

\rowcolor{gray!10}
 & \textbf{Dense} & \textbf{Font} & \textbf{Style} & \textbf{Composition} & \textbf{Intention} & \\
 \rowcolor{gray!10}
 \cline{2-6}
 \rowcolor{gray!10}
\multirow{-2}{*}{\textbf{Model}} 
& $PS$ $\uparrow$ & $R$ $\uparrow$ & $S$ $\uparrow$ & $PS$ $\uparrow$ & $PS$ $\uparrow$ 
& \multirow{-2}{*}{\textbf{Average} $\uparrow$} \\

\midrule
Seedream-4.0                & 0.618 & 0.342 & 0.591 & 0.848 & 0.645 & 0.609 \\
Gemini-2.5-Flash-Image      & \textbf{0.622} & \textbf{0.391} & 0.590 & \textbf{0.866} & 0.663 & \textbf{0.626} \\
GPT-Image-1                 & 0.508 & 0.299 & \textbf{0.633} & 0.856 & \textbf{0.670} & 0.593 \\
Qwen-Image                  & 0.464 & 0.286 & 0.620 & 0.801 & 0.589 & 0.552 \\
\bottomrule
\end{tabular}%
}
\label{tab:image_gen_results}
\vspace{-5mm}
\end{table}

\begin{figure*}[t]
    \centering
    \includegraphics[width=1\textwidth]{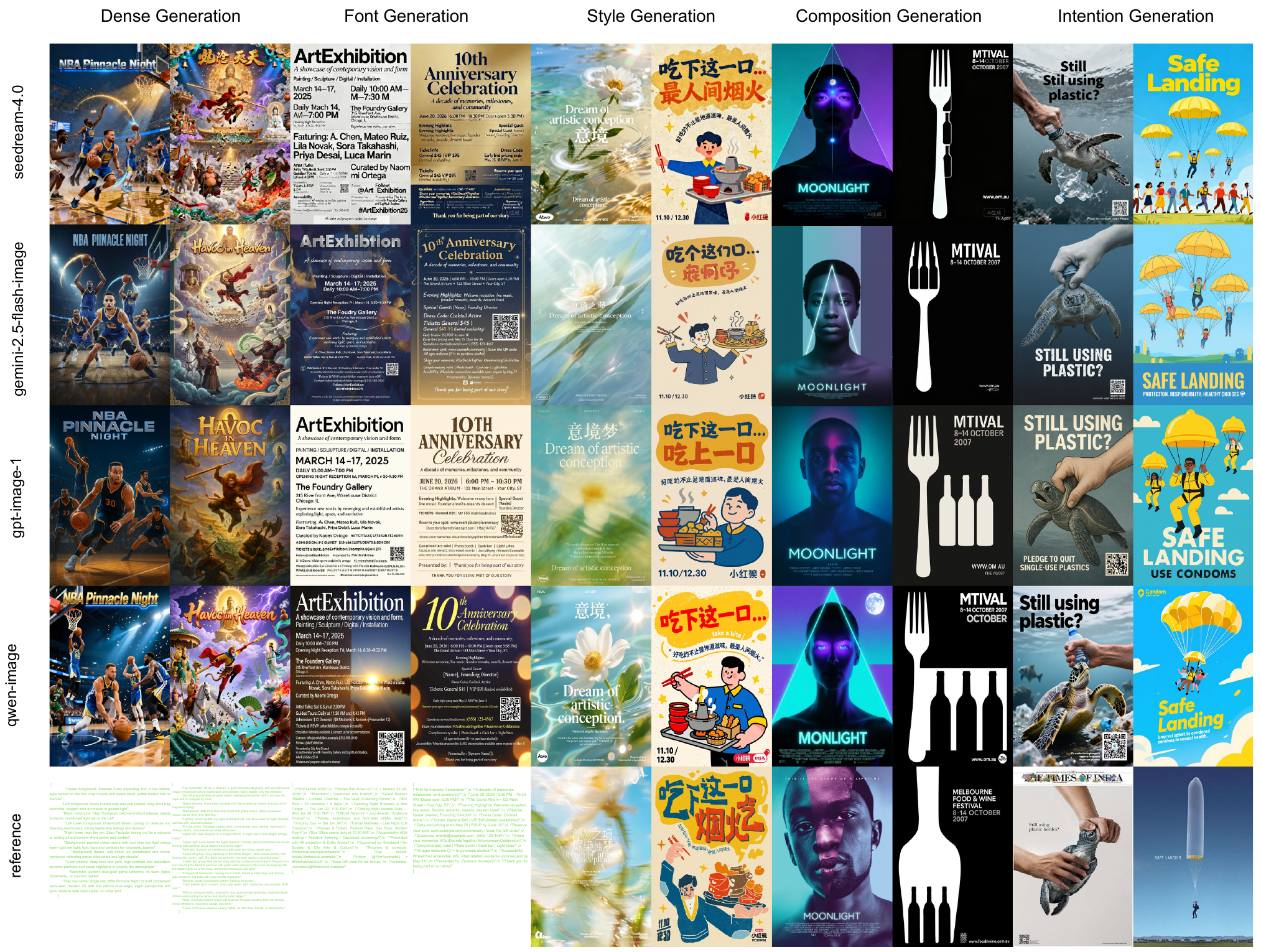}
    \vspace{-7mm}
    \caption{Qualitative comparison of four models on five generation tasks.}
    \vspace{-5mm}
    \label{fig:generation}
\end{figure*}

\textbf{Qualitative Results.} The Dense Generation task targets crowded scenes with many subjects, rich local detail, and strong interactions, emphasizing facial fidelity, hands, textures, and clean edges (Fig.~\ref{fig:generation}, cols.~1--2). Gemini-2.5-Flash-Image delivers the most stable faces and skin; \textit{GPT-Image-1} is close behind; Qwen-Image often over-brightens, yielding unnatural texture; Seedream-4.0 is weakest, with frequent collapses and distortions. In cartoons, Gemini-2.5-Flash-Image preserves crisp edges and stylistic consistency; GPT-Image-1 drifts under complexity. 
For Font Generation (Fig.~\ref{fig:generation}, cols.~3--4), Seedream-4.0 explores the boldest styles; Gemini-2.5-Flash-Image also varies fonts; GPT-Image-1  and Qwen-Image prefer conservative, upright forms. Seedream-4.0 has the lowest text sharpness and breaks strokes under layered effects. In Style Generation, all models broadly match targets, but dispersion style (Fig.~\ref{fig:generation}, col.~5) remains difficult; Qwen-Image stays overly sharp and saturated, and often misses vintage tone; Gemini-2.5-Flash-Image and GPT-Image-1  skew more vintage, with GPT-Image-1  leaning comic and toward colored backgrounds; Seedream-4.0 tends toward naturalistic renderings with moderate completeness; Gemini-2.5-Flash-Image shows a Western bias. Compared with ground-truth designs (Fig.~\ref{fig:generation}, col.~6), all converge to safe, low-creativity styles.
In Composition Task, Gemini-2.5-Flash-Image shows stronger grasp of nested structures, yet advanced layouts---intentional misalignment, whitespace, figure--ground play, cutouts---generally fail (Fig.~\ref{fig:generation}, col.~7); fusion/collage exhibits seams and fragmentation; complex spatial relations with whitespace lead to poor saliency and attention allocation. Results in Intention Task reveal that existing models manage surface associations but rarely convey deeper concepts. All in all, end-to-end image synthesis remains insufficient for highly constrained poster design; a full pipeline---\emph{understand} $\rightarrow$ \emph{ideate} $\rightarrow$ \emph{compose} $\rightarrow$ \emph{generate}---is needed to meet high-level design constraints.

\begin{figure}[t]
    \centering
    \includegraphics[width=1\linewidth]{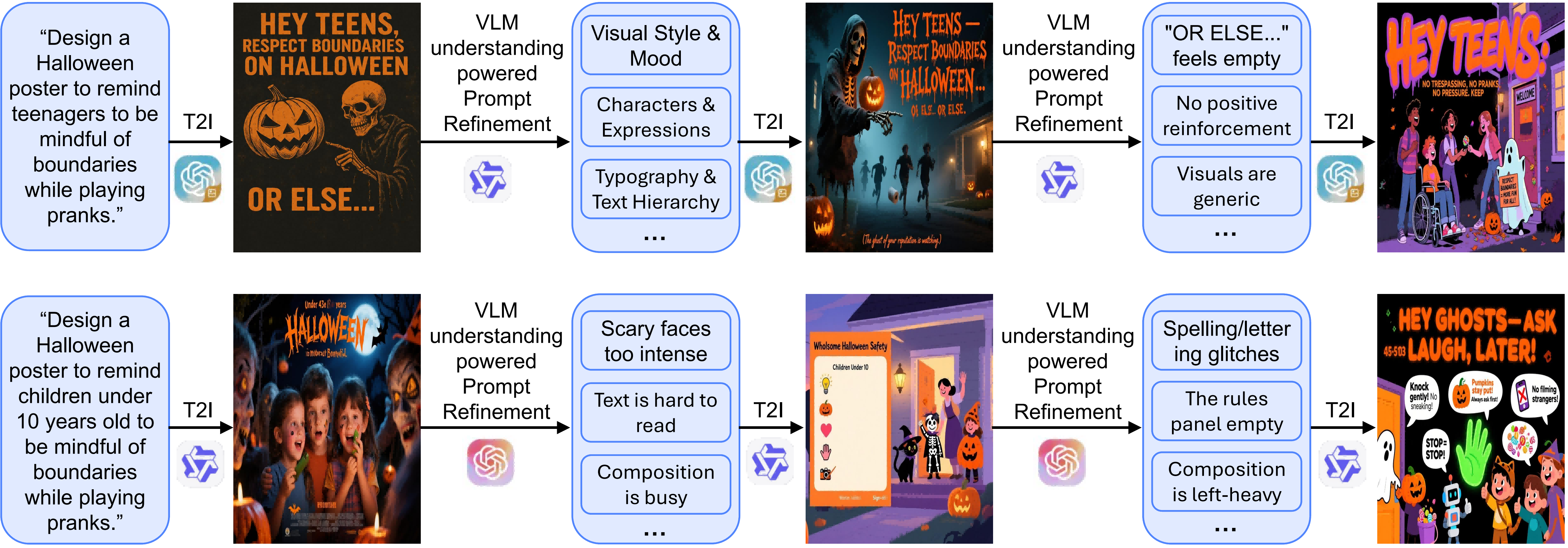}
    \vspace{-6mm}
    \caption{Qualitative comparison of model outputs over supervision-guided iterations.}
    \vspace{-2mm}
    \label{fig:data-format}
    
\end{figure}

\textbf{Understanding and Generation.} We investigate how improved understanding benefits generation and validate the soundness of our benchmark. To this end, we design an iterative setup where a VLM supervises a T2I generator. We simulate a realistic use case: the input is a vague, brief requirement as the prompt. A T2I model first produces a poster. We then feed the generated poster and the original prompt into a VLM. Based on its interpretation of the poster, the VLM diagnoses issues and returns a revised prompt, which is fed back to the generator for re-generation. Notably, the VLM only receives the image and the original prompt; the enhanced prompt without any human or domain-specific design hints.
By iterating this loop, we directly observe how understanding affects poster quality. As shown in Fig.~\ref{fig:data-format}, the two T2I models are GPT-Image-1 and Qwen-Image, and the two VLMs are GPT-5 and Qwen3-VL. From the first T2I outputs under ambiguous requirements, both models capture the main content, but GPT-Image exhibits stronger intent understanding than Qwen-Image. It conveys a cautionary message to young people about excessive play, whereas Qwen-Image attends to surface keywords such as “Halloween” and “children,” without deeper inference. After the first round of VLM-guided analysis, the VLM flags issues in visual style, mood, and typography. The regenerated results from the revised prompt address these issues and improve overall poster quality. A second iteration yields further gains, confirming that stronger understanding improves generation.
At round 0 (no understanding intervention), the models tend to produce visually appealing images that neglect efficient communication. After VLM intervention, the system begins to prioritize effective information transmission. The clearest evidence is the adjustment of emotional tone to match communication goals. Without intervention, round-0 results often include elements unsuitable for the audience, such as frightening skulls or uncanny faces. With deeper understanding, these issues are corrected. These findings suggest a practical paradigm for poster generation: iteratively coupling understanding and generation—without human design hints.

\begin{table}[t]
\centering
\caption{Average scores across understanding tasks. The \textit{average} is computed by aggregating only these positively correlated scores.}
\vspace{-2mm}
\renewcommand{\arraystretch}{1.2}
\resizebox{0.48\textwidth}{!}{%
\begin{tabular}{|l|c|c|c|c|c|c|}
\toprule
\rowcolor{gray!10}\textbf{Model} & \textbf{OCR} & \textbf{Font} & \textbf{Layout} & \textbf{Understanding} & \textbf{Overall Rating} & \textbf{Average} \\
\midrule
GPT-5                 & 0.838 & \textbf{0.615} & 0.511 & 0.801 & 0.347 & 0.622\\
Claude-Sonnet-4.5     & 0.838 & 0.603 & 0.453 & 0.727 & 0.384 & 0.601\\
Gemini-2.5-Pro        & \textbf{0.855} & 0.606 & \textbf{0.571} & \textbf{0.806} & 0.399 & \textbf{0.647}\\
Grok-4-fast           & 0.475 & 0.420 & 0.313 & 0.682 & 0.465 & 0.471\\
\hline
MiniCPM-V-4.5         & 0.838 & 0.421 & 0.325 & 0.652 & 0.095 & 0.466\\
Gemma-3n-e4b-it       & 0.744 & 0.381 & 0.287 & 0.538 & \textbf{0.483} & 0.486\\
Qwen3-VL-4B  & 0.835 & 0.480 & 0.427 & 0.726 & 0.172 & 0.538\\
Qwen3-VL-8B  & 0.841 & 0.445 & 0.526 & 0.668 & 0.237 & 0.543\\
\bottomrule
\end{tabular}%
}
\vspace{-5mm}
\label{tab:avg_scores}
\end{table}

\textbf{Overall Evaluation.}  Finally, we aggregate model performance into an overall score. For each understanding tasks, we first average all positively oriented metrics to obtain group-wise scores for OCR, Font, Layout, Understanding, and Overall Rating. We then take the mean of these five group scores to derive the final score for each model, as reported in Tab.~\ref{tab:avg_scores}. The three proprietary models GPT-5, Claude-Sonnet-4.5, and Gemini-2.5-Pro exhibit stronger poster understanding capabilities than the open-source models. Combined with the generation results in Tab.~\ref{tab:image_gen_results}, this suggests that Gemini-2.5 achieves a consistently strong balance between understanding and generation.

\section{Conclusion}
PosterIQ delivers a rigorous benchmark for poster design. By decoupling core capabilities, our tasks expose where current MLLMs and generators succeed and where they fail, from saliency control to intention communication. Empirical studies reveal that frontier models excel at high-level reasoning yet remain insensitive as automatic raters and struggle with composition-aware synthesis and typographic diversity. We hope this work catalyzes multimodal intelligence and guides future models toward fidelity to design constraints, communicative intent, and creativity in real-world poster scenes.
\clearpage

\section*{Acknowledgement}
The work described in this paper was substantially supported by a grant from the Research Grants Council of the Hong Kong Special Administrative Region, China (Project No. PolyU/RGC Project  25211424) and partially supported by a grant from PolyU University Start-Up Fund (Project No. P0047675).

{
    \small
    \bibliographystyle{ieeenat_fullname}
    \bibliography{main}
}
\clearpage
\clearpage
\appendix
\setcounter{page}{1}
\maketitlesupplementary

We first present the statistical details of PosterIQ, followed by a description of how we obtain the evaluation results for each task. To validate the automatic evaluation, we also conduct a human evaluation and provide the annotator guideline used to construct the benchmark. Finally, we provide visual examples for each task to aid understanding.

\section{Benchmark Statistics}
\begin{figure}[t]
    \centering

    \includegraphics[width=0.8\linewidth]{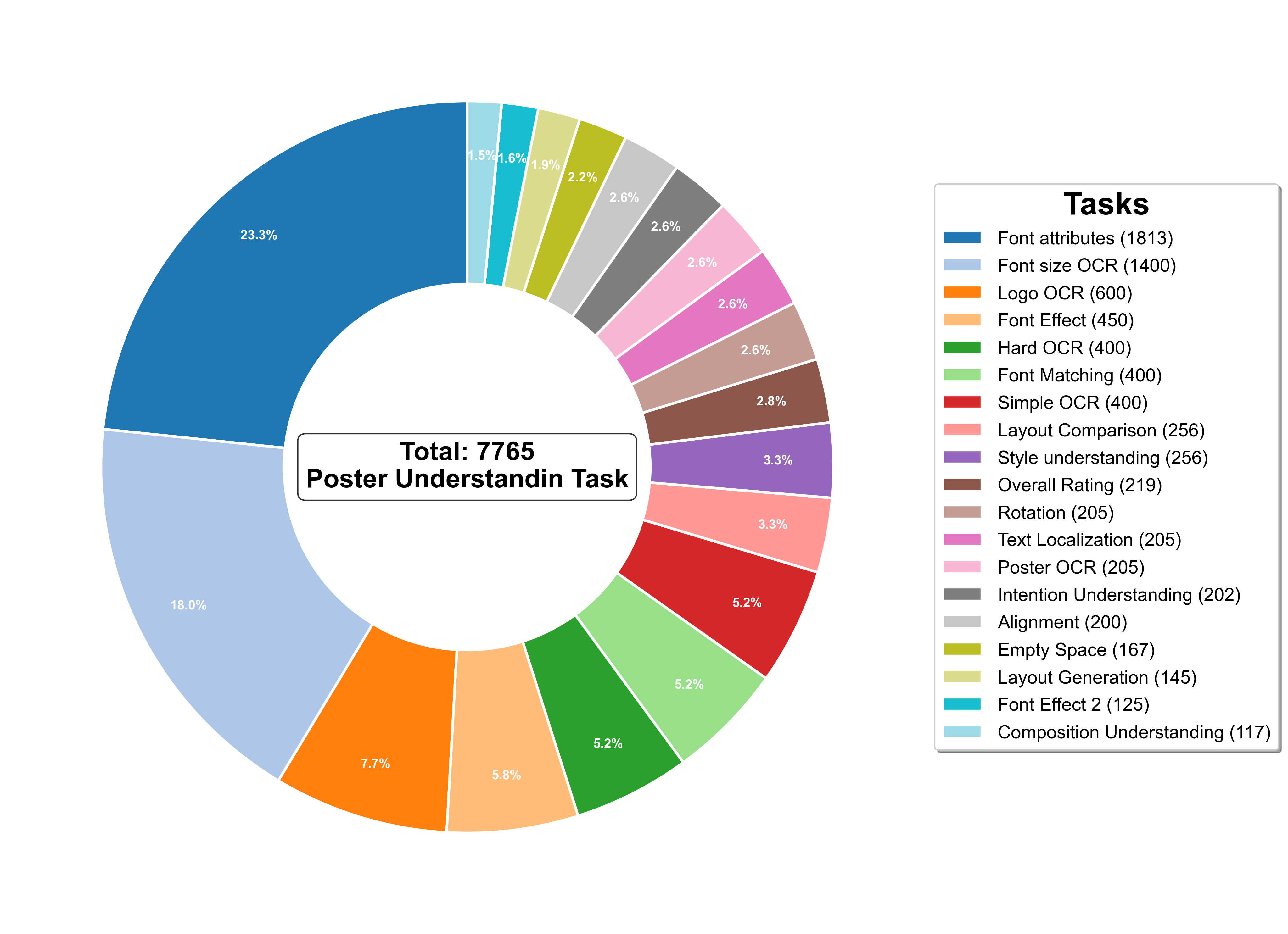}

    \vspace{-1em}

    \includegraphics[width=0.8\linewidth]{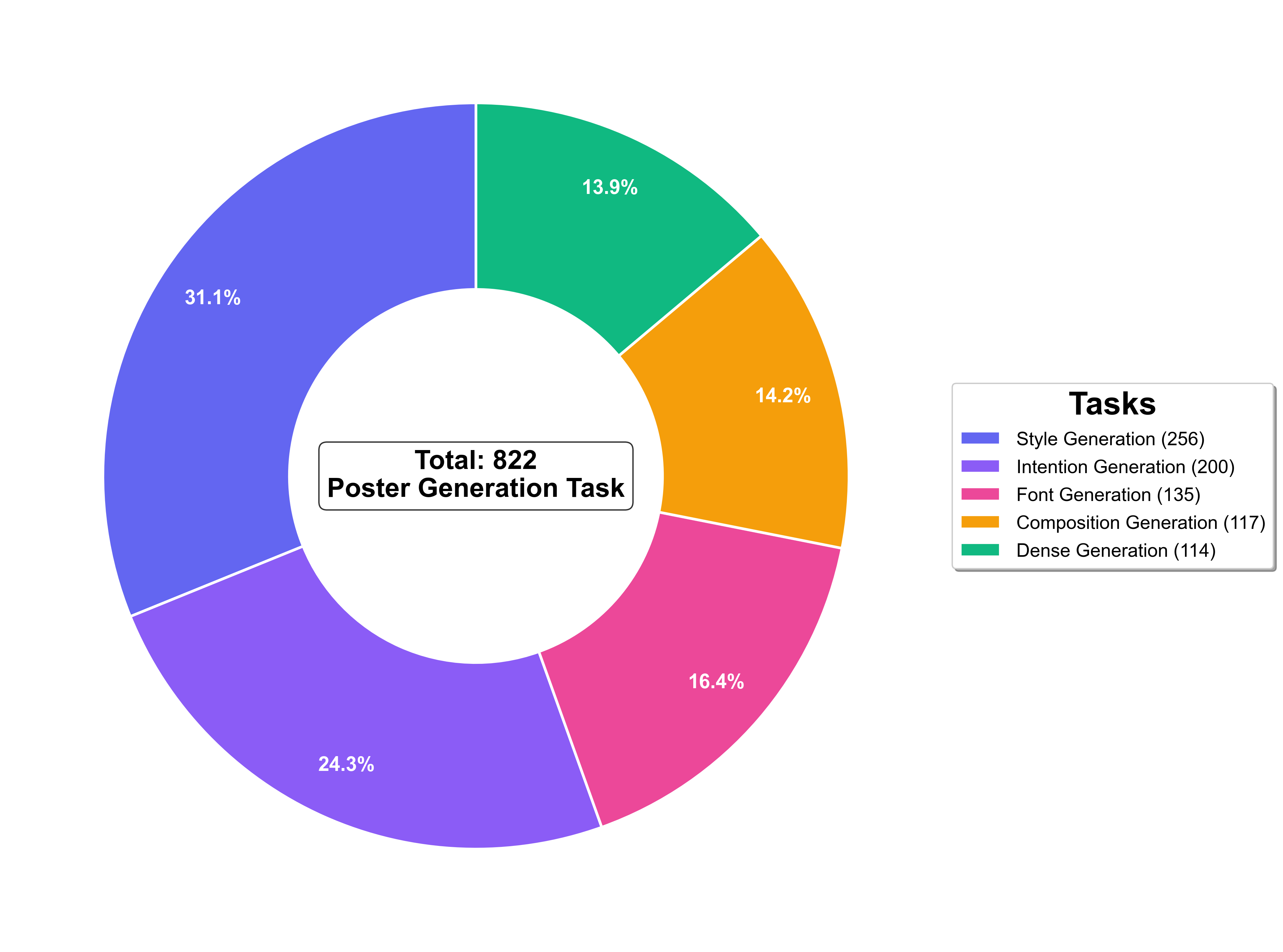}

    \caption{Benchmark statistics for understanding tasks (top) and generation tasks (bottom).}
    \label{fig:benchmark_stats}
\end{figure}

Figure~\ref{fig:benchmark_stats} summarizes the data distribution of our benchmark.
For the \textbf{understanding} part (top), the dataset contains 7{,}765 items in total, with font-related tasks taking the largest share: \emph{Font Attributes} (1{,}813, 23.3\%) and \emph{Font Size OCR} (1{,}400, 18.0\%) together account for over 40\% of all instances. 
OCR and layout–related tasks, including \emph{Logo OCR}, \emph{Poster OCR}, \emph{Simple/Hard OCR}, \emph{Text Localization}, \emph{Layout Comparison}, \emph{Empty Space}, and \emph{Layout Generation}, form the bulk of the remaining samples, while \emph{Style Understanding}, \emph{Composition Understanding}, \emph{Intention Understanding}, and \emph{Overall Rating} provide higher-level assessments of visual design and semantics.

For the \textbf{generation} part (bottom), the 822 instances are evenly distributed: \emph{Style Generation} (256, 31.1\%) and \emph{Intention Generation} (200, 24.3\%) dominate the set, whereas \emph{Font Generation} (135), \emph{Composition Generation} (117), and \emph{Dense Generation} (114) each contribute roughly 14–16\% of the total, ensuring balanced coverage across different aspects of poster synthesis.

\section{Task Evaluation}
\noindent\textbf{OCR Accuracy (Text Instance Level):}
For \textbf{logo OCR} and \textbf{poster OCR}, we evaluate accuracy at the text-instance level. 
Invisible characters (e.g., spaces and line breaks) are stripped from both prediction and ground truth, and an instance is counted as correct only under exact match. 
The overall accuracy (\textit{AC}) is then given by the average proportion of correctly recognized text instances:

\begin{equation}
\textit{AC} = \frac{1}{N} \sum_{i=1}^N \mathbbm{1}\left( \hat{T}_i = T_i \right),
\end{equation}

where \(N\) is the total number of text instances, \(\hat{T}_i\) is the predicted text for instance \(i\), and \(T_i\) is the corresponding ground-truth text. 
The function \(\mathbbm{1}(\cdot)\) is an indicator that returns 1 if the condition holds and 0 otherwise. 
Each logo is counted as a single text instance, while a poster can contain multiple text instances.

\noindent\textbf{Word-level Recall for Robust OCR:}
This metric is used for the synthetic OCR tasks: \textbf{Simple OCR}, \textbf{Hard OCR}, and \textbf{Font-Size OCR}. 
Given a ground-truth text string, we split it into consecutive units \( g \), each consisting of five characters. 
For each unit \( g \), we check whether it appears as a substring in the model-predicted long text \( \textit{output} \). 
This normalization makes scores comparable across texts of different lengths. 
The \textit{word-level recall} ($WR$) is then defined as the proportion of ground-truth units that are recovered in the prediction:

\begin{equation}
\textit{$WR$} = \frac{\sum_{g \in G} \mathbbm{1}\left( g \subseteq output \right)}{|G|},
\end{equation}

where \( G \) denotes the set of all five-character units extracted from the ground-truth text, and \(\mathbbm{1}(\cdot)\) is an indicator function that returns 1 when the condition is satisfied and 0 otherwise. Before segmentation, spaces and escape characters are stripped, and each word is guaranteed to appear at most once in a given image.

We define \(\Delta\) as the difference between the \textit{WR} scores on \textbf{Simple OCR} and \textbf{Hard OCR}. 
This gap reflects the robustness of the model to noise: a larger \(\Delta\) indicates higher sensitivity to background clutter, missing context, rotation, and other perturbations.

For the \textbf{Font-Size OCR} task, we additionally report the standard deviation \(\textit{Std}\) of the \textit{WR} scores across 14 different font sizes. 
This metric captures the robustness of the model’s OCR performance with respect to changes in font size: lower \(\textit{Std}\) indicates more stable recognition across scales.

\noindent\textbf{K-Option Scoring for Multiple-Choice Tasks:} For tasks where the model predicts a label from a finite set of options, we use multiple-choice style metrics. 
This applies to the font-related tasks (\textbf{Font Matching}, \textbf{Font Attributes}, and \textbf{Font Effects}), where we report \emph{Score}, \emph{Effect Score}, and \emph{Color Score}; 
to the \textbf{Style Understanding} task, where we report a style classification \emph{Score}; 
to the \textbf{Text Position} task, where \emph{Alignment} and \emph{Rotation} are cast as discrete choices; 
and to \textbf{Layout Comparison}, where we also use a multiple-choice \emph{Score}.
For multiple-choice tasks with \( k \) answer options, let the model's accuracy be \( a \), where \( a \in [0,1] \). The scoring formula normalizes the score such that random guessing results in a score of zero, and perfect accuracy results in a score of one:

\begin{equation}
\text{Score} = \max \left( 0, \frac{k \cdot a - 1}{k - 1} \right).
\end{equation}

Under random guessing, the expected accuracy is \( \frac{1}{k} \), which maps to a score of zero in our formulation. 
When the model attains perfect accuracy \( a = 1 \), the score reaches one. 

This scoring scheme is used for the font-related tasks, the text positioning task, and the layout comparison task.

\noindent\textbf{Bbox-Related Metrics:}
For the \textbf{Text Localization} task, ground-truth bounding boxes are first sorted in descending order by area. 
We then evaluate the average Intersection over Union (IoU) over the top-\(n\) predicted boxes:
\begin{equation}
\text{IoU} = \frac{1}{n} \sum_{j=1}^n 
\frac{\lvert B_j^{\text{pred}} \cap B_j^{\text{gt}} \rvert}
     {\lvert B_j^{\text{pred}} \cup B_j^{\text{gt}} \rvert},
\end{equation}
where \(B_j^{\text{pred}}\) and \(B_j^{\text{gt}}\) denote the predicted and ground-truth boxes for the \(j\)-th instance.

To further assess prompt-following behavior, we examine how well the number of predicted boxes matches the number of queried text instances. 
For each sample, if the model predicts fewer boxes than requested, we compute the recall as the ratio between the number of predicted boxes and the number of queried objects. 
If it predicts more boxes than requested, we assign a recall of 1. 
The final recall score is the average over all samples:
\begin{equation}
\text{Recall Rate} = \frac{1}{N} \sum_{i=1}^N 
\begin{cases}
\frac{n_i^{\text{pred}}}{n_i^{\text{query}}}, & \text{if } n_i^{\text{pred}} \le n_i^{\text{query}}, \\[4pt]
1, & \text{otherwise},
\end{cases}
\end{equation}
where \(N\) is the total number of evaluation samples, \(n_i^{\text{pred}}\) is the number of predicted boxes for sample \(i\), and \(n_i^{\text{query}}\) is the number of query objects specified in the prompt.

In the \textbf{Layout Generation} task, there is no uniquely correct placement of layout boxes, so standard IoU-based matching is not directly applicable. 
Instead, we evaluate the predicted layout by comparing the relative positions and areas of predicted boxes with those of the ground truth. 
Higher-quality layouts exhibit smaller \emph{Center Bias} and an \emph{Area Ratio} closer to 1.

\textit{Center Bias} quantifies the normalized Euclidean distance between the centers of the predicted and ground-truth boxes:
\begin{equation}
\text{Center Bias} = \frac{1}{N} \sum_{i=1}^N \bigl\| C_i^{\text{pred}} - C_i^{\text{gt}} \bigr\|_2,
\end{equation}
where \(C_i^{\text{pred}}\) and \(C_i^{\text{gt}}\) are the normalized center coordinates of the predicted and ground-truth box for the \(i\)-th element.

\textit{Area Ratio} measures how similar the box areas are by taking the ratio between the smaller and larger area:
\begin{equation}
\text{Area Ratio} = \frac{1}{N} \sum_{i=1}^N 
\frac{\min\!\bigl(A_i^{\text{pred}}, A_i^{\text{gt}}\bigr)}
     {\max\!\bigl(A_i^{\text{pred}}, A_i^{\text{gt}}\bigr)},
\end{equation}
where \(A_i^{\text{pred}}\) and \(A_i^{\text{gt}}\) denote the areas of the predicted and ground-truth boxes, respectively.

\noindent\textbf{Empty-Space Evaluation.}
In the \textbf{Empty-Space} task, both the ground truth and the model output are represented as sets of patch IDs. 
The ground truth set corresponds to the patches annotated as suitable empty regions, and the model is asked to predict a set of patch IDs for placing new content. 
We first measure the agreement between these sets using Intersection over Union (IoU):
\begin{equation}
\text{Patch IoU} = \frac{1}{N} \sum_{i=1}^N 
\frac{\lvert \mathcal{P}_{\text{pred}} \cap \mathcal{P}_{\text{gt}} \rvert}
     {\lvert \mathcal{P}_{\text{pred}} \cup \mathcal{P}_{\text{gt}} \rvert},
\end{equation}
where \(\mathcal{P}_{\text{pred}}\) and \(\mathcal{P}_{\text{gt}}\) denote the predicted and ground-truth patch ID sets, respectively.

\textit{Match Accuracy.}
The prompt also specifies how many patch IDs should be returned. 
We therefore evaluate prompt-following behavior by checking whether the predicted set size matches the requested size. 
Match Accuracy is defined as the proportion of samples that satisfy this constraint:
\begin{equation}
\text{Match Accuracy} = \frac{1}{N} \sum_{i=1}^N 
\mathbbm{1}\bigl( \lvert \mathcal{P}_{\text{pred}} \rvert = \lvert \mathcal{P}_{\text{gt}} \rvert \bigr),
\end{equation}
where \(N\) is the number of evaluation samples, \(\mathcal{P}_{\text{pred}}\) is the predicted patch set for sample \(i\), 
\(\lvert \mathcal{P}_{\text{gt}} \rvert\) is the number of patch IDs requested in the prompt, 
and \(\mathbbm{1}(\cdot)\) is an indicator function that returns 1 if the condition holds and 0 otherwise.

\noindent\textbf{Point Score for Advanced Understanding Metrics:}
For \textbf{intention understanding}, each creative advertisement poster is paired with a set of manually annotated key elements that summarize the intended semantic or conceptual message of the design. To evaluate whether an MLLM can correctly capture and verbalize these elements, we use GPT-5 as an automatic judge. Given a model-generated caption, the judge checks for each key element whether it is correctly identified and explicitly mentioned. For a given poster, the prediction is labeled \texttt{Yes} if all annotated key points are covered, and \texttt{No} otherwise. The resulting \emph{Point Score} is defined as the fraction of posters judged as \texttt{Yes}:
\begin{equation}
\text{Point Score} = \frac{N_{\text{Yes}}}{N_{\text{Total}}},
\label{eq:point_score}
\end{equation}
where \(N_{\text{Yes}}\) is the number of posters whose model-generated captions successfully cover all key points, and \(N_{\text{Total}}\) is the total number of evaluated posters.

For \textbf{Composition Understanding}, we adopt the same Point Score metric.

\noindent\textbf{Overall Rating Metric:}
In the \textbf{Overall Rating} task, both humans and MLLMs assign a quality score in the range 0--10 for each poster. 
We first normalize human and model scores to have zero mean, and then measure their agreement via cosine similarity. 
Formally, let \(\mathbf{h} \in \mathbb{R}^N\) and \(\mathbf{m} \in \mathbb{R}^N\) denote the human and model score vectors over \(N\) posters. 
We compute the zero-mean versions
\begin{equation}
\tilde{\mathbf{h}} = \mathbf{h} - \bar{h}\mathbf{1}, \quad
\tilde{\mathbf{m}} = \mathbf{m} - \bar{m}\mathbf{1},
\end{equation}
where \(\bar{h}\) and \(\bar{m}\) are the mean human and model scores, and \(\mathbf{1}\) is an all-ones vector. 
The final metric is the cosine similarity between the two normalized vectors:
\begin{equation}
\text{Overall Rating} = 
\frac{\tilde{\mathbf{h}}^\top \tilde{\mathbf{m}}}
     {\|\tilde{\mathbf{h}}\|_2 \,\|\tilde{\mathbf{m}}\|_2}.
\end{equation}

\noindent\textbf{Point Score for Poster Generation:}
For the \textbf{Dense Generation}, \textbf{Composition Generation}, and \textbf{Intention Generation} tasks, we use the \emph{Point Score} to evaluate whether the generated image covers all required key elements. 
Each generated poster is associated with multiple checkpoints (e.g., required objects, layout cues, or semantic intentions), and a MLLM judge determines for each checkpoint whether it is correctly realized in the image.

Concretely, we use GPT-5 as the automatic judge for Dense Generation and Intention Generation, and Gemini-2.5-Pro as the judge for \textbf{Composition Generation}. 
The Point Score is then computed as in Eq.~\eqref{eq:point_score}, i.e., as the fraction of images whose generated content is judged to cover all annotated key points.

\noindent\textbf{Score for Style Generation.}
In the \textbf{Style Generation} task, the generative model is instructed (via a textual prompt) to produce a poster in a specified target style. 
A MLLM is then asked to classify the generated poster into one of the predefined style labels. 
We compare the predicted style label with the ground-truth target label and compute a style generation score as the accuracy over all evaluated samples:

\begin{equation}
\text{Style Score} = \frac{1}{N} \sum_{i=1}^{N} 
\mathbbm{1}\bigl( \hat{s}_i = s_i^{\text{gt}} \bigr),
\end{equation}

where \(N\) is the number of generated posters, \(\hat{s}_i\) is the style label predicted by the MLLM (GPT-5) for the \(i\)-th poster, 
\(s_i^{\text{gt}}\) is the corresponding ground-truth style label, and \(\mathbbm{1}(\cdot)\) is the indicator function.

\noindent\textbf{Font Richness for Font Generation.}
In the \textbf{Font Generation} task, the generative model is prompted to produce posters with diverse typography, where the prompt explicitly specifies the target text and encourages the use of varied font styles. 
After generation, we ask an MLLM with strong font understanding ability (GPT-5) to describe the typography of each poster using a fixed vocabulary of font attributes (e.g., \emph{modern}, \emph{playful}, \emph{serif}, \emph{italic}, etc.).

Let \(\mathcal{A}\) be the set of all font attributes (e.g., the 37 attributes in our implementation), and \(|\mathcal{A}| = M\). 
For a batch of \(N\) generated posters, we define a binary indicator \(x_{i,a} \in \{0,1\}\) that equals 1 if GPT-5 assigns attribute \(a \in \mathcal{A}\) to the \(i\)-th poster, and 0 otherwise. 
For each attribute \(a\), we first compute its coverage ratio over the batch:
\begin{equation}
R_a = \frac{1}{N} \sum_{i=1}^{N} x_{i,a},
\end{equation}
which measures how frequently attribute \(a\) appears across generated posters.

The overall \emph{Font Richness Score} is then defined as the average coverage ratio over all attributes:
\begin{equation}
\text{Richness} = \frac{1}{M} \sum_{a \in \mathcal{A}} R_a
= \frac{1}{NM} \sum_{a \in \mathcal{A}} \sum_{i=1}^{N} x_{i,a}.
\end{equation}
Intuitively, this metric reflects how widely the generator explores the font attribute space: higher values indicate that a broader range of font attributes is realized across the generated posters.

\section{Human Evaluation}
To verify the reliability of our automatic evaluation, we conduct a series of human studies on both understanding and generation tasks. 
For several understanding tasks that rely on LLM-based textual judgments (e.g., \textbf{Composition Understanding} and \textbf{Intention Understanding}), we compare the decisions of the automatic judge with those of human annotators. 
On a subset, the agreement between the LLM judge and human evaluation reaches approximately 92\%, indicating that our LLM-as-judge protocol is largely consistent with human judgments.

For the \textbf{Generation} tasks, we employ MLLMs to repeatedly assess the quality and faithfulness of generated images. 
Specifically, in \textbf{Dense Generation}, \textbf{Composition Generation}, and \textbf{Intention Generation}, the judge verifies whether multiple key pieces of information are correctly rendered in the image, while in \textbf{Font Generation} and \textbf{Style Generation}, the judge directly assigns font or style labels to each poster. 
To validate these automatic scores, we randomly sample 30 generated images per task and obtain human ratings under the same criteria. 
We observe that the relative ranking of generative models remains largely consistent across different MLLM judges and human annotators, suggesting that our automatic evaluation provides a stable and trustworthy proxy for human assessment.

\section{Annotator Guideline}

We adopt a multi–stage pipeline for data collection and annotation. 
First, we gather poster images from free sources that explicitly permit research use. 
All raw images are manually cleaned to remove samples with blurry content, severe artifacts, or copyright concerns. 
For OCR-related understanding tasks, we rely on reliable digital sources as ground-truth text. 
Human annotation is mainly required for layout-related tasks, advanced understanding tasks, and the overall rating task.  

For each such task, at least three expert annotators independently label every sample. 
A senior annotator (the \emph{leader}) then cross-checks all submissions, resolves disagreements, and filters out ambiguous cases, retaining only samples with high inter-annotator agreement. 
Below we summarize the concrete annotation guidelines for representative tasks.

\noindent\textbf{Empty Space Task.}
We begin from partially edited poster designs, where some design elements have been intentionally removed from the original PSD files.  
The resulting posters are rendered with an overlaid grid, and the grid patch indices are visible to annotators.  
Each poster is sent to three annotators with the following instruction:
\emph{“This is an unfinished poster. New design elements need to be added. Please identify all patch IDs that you consider suitable empty regions for placing new content.”}  
The leader aggregates the proposed patch sets and retains only those samples whose recommended regions achieve more than 90\% agreement across annotators, making a final decision when minor discrepancies occur.

\noindent\textbf{Composition Understanding Task.}
We collect posters that exhibit strong visual reconstruction or structural composition (e.g., displacement, nesting, segmentation).  
Each poster is distributed to multiple annotators with the instruction:
\emph{“Using concise natural language, list the visual design techniques used in this poster (such as displacement, nesting, segmentation, extension, focus, mirroring, cut-out, arrangement, etc.). Describe only the necessary composition cues in bullet points.”}  
The leader reviews and consolidates all descriptions, and keeps only those posters for which different annotators provide highly consistent composition cues.

\noindent\textbf{Intention Understanding Task.}
We curate posters that contain clear visual metaphors or conceptual designs.  
Each poster is assigned to several annotators with the instruction:
\emph{“First, carefully read the content in the poster. Then, search for the original source or explanation of this poster online. If the external explanation aligns with your own understanding, keep this sample and decompose its core metaphor or concept into several key pieces of information. If the external explanation conflicts with your interpretation, discard this sample.”}  
The leader then collects and refines the key-intention annotations, merging overlapping items and removing noisy or inconsistent samples.

\noindent\textbf{Overall Rating Task.}
For the overall quality assessment, we distribute each poster to multiple annotators with the instruction:
\emph{“Please rate the overall design quality of this poster on a scale from 0 to 10, where 0 is the worst and 10 is the best. Consider font properties, layout, textual communication, and creative concept in your score.”}
Because different annotators may use different scoring ranges, we first standardize their score distributions (zero-mean and variance normalization), and then discard posters whose inter-annotator score range exceeds a predefined threshold.  
The remaining posters, which exhibit high rating consistency, are averaged to obtain a stable ground-truth score used in our benchmark.

\clearpage
\onecolumn 

\section{Task Illustration}
\vspace{-6mm}
\begin{figure}[H]
    \centering
    \includegraphics[width=1\textwidth]{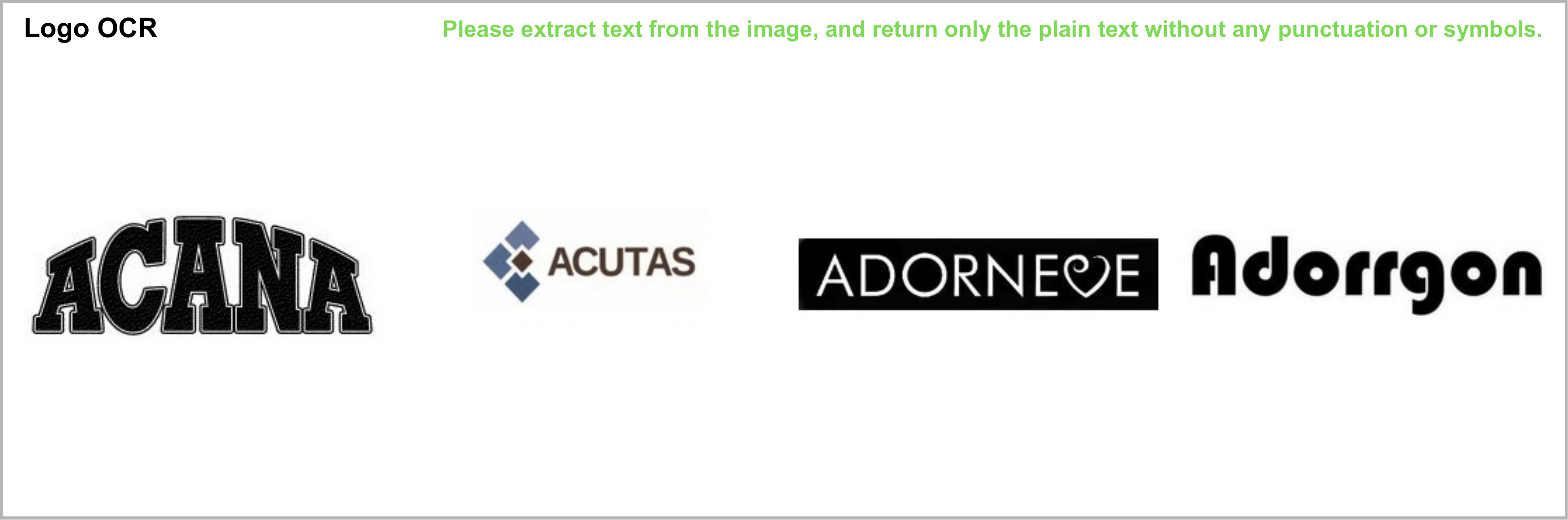}
\end{figure}
\vspace{-6mm}
\begin{figure}[H]
    \centering
    \includegraphics[width=1\textwidth]{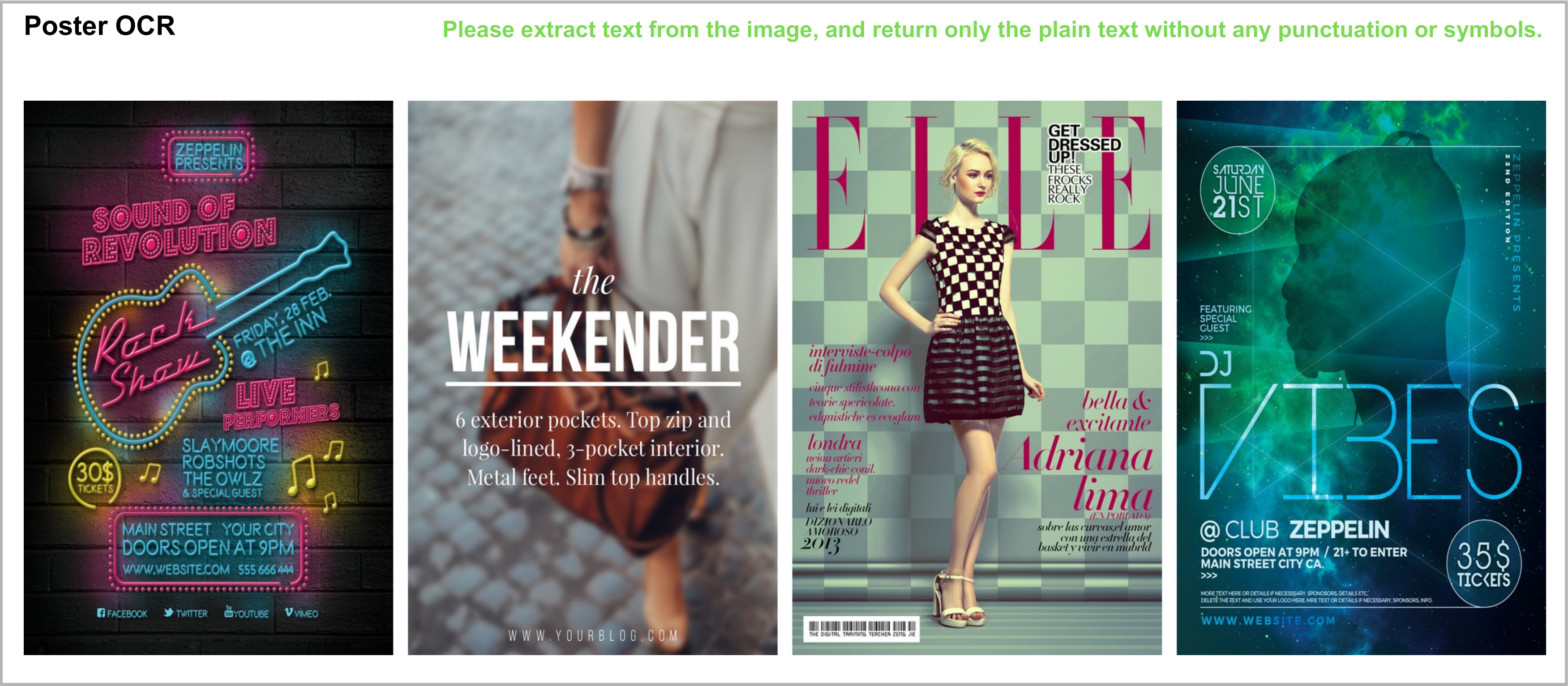}
\end{figure}
\vspace{-6mm}
\begin{figure}[H]
    \centering
    \includegraphics[width=1\textwidth]{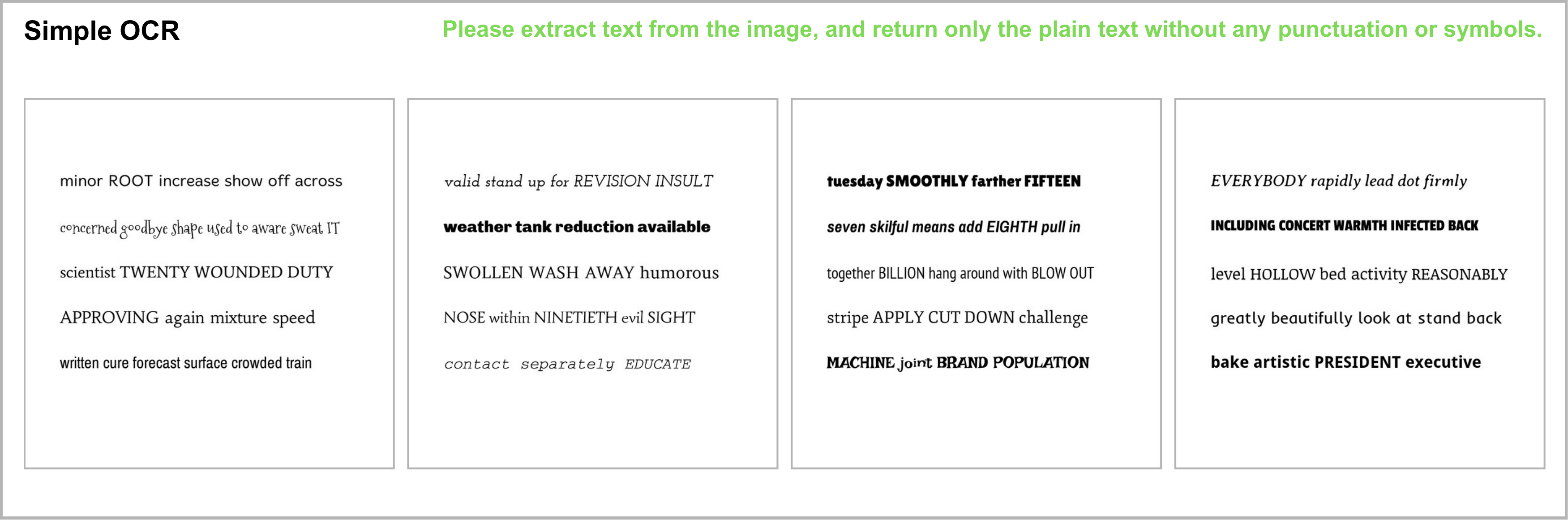}
\end{figure}
\vspace{-6mm}
\begin{figure}[H]
    \centering
    \includegraphics[width=1\textwidth]{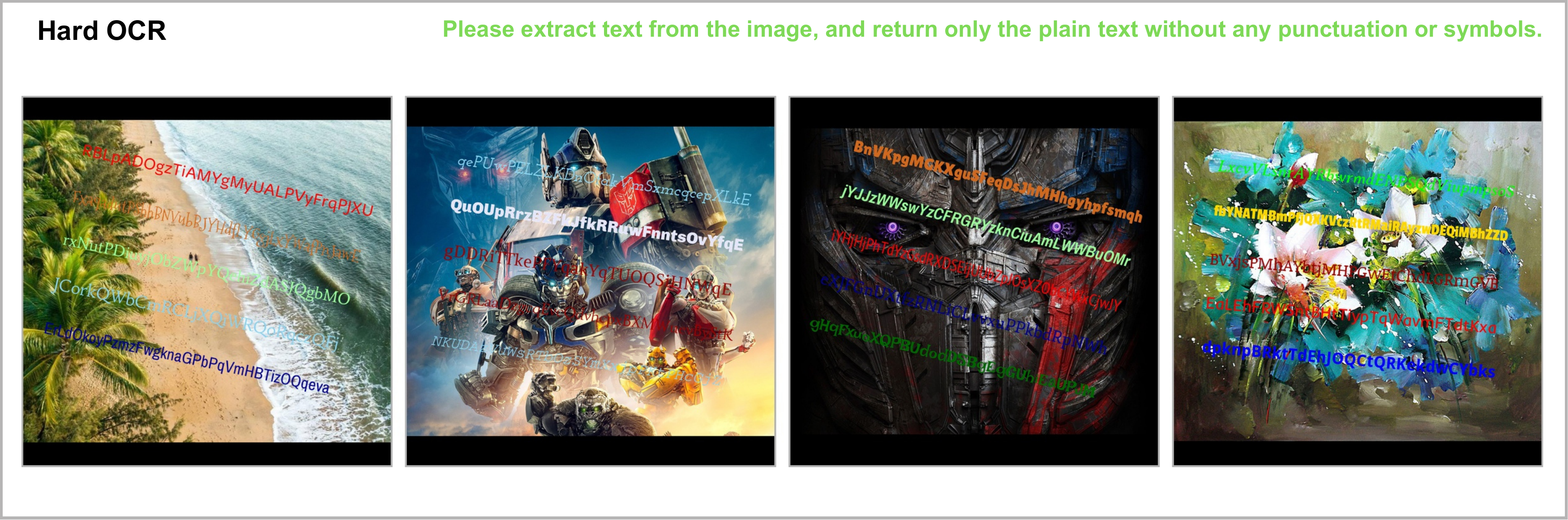}
\end{figure}
\vspace{-6mm}
\begin{figure}[H]
    \centering
    \includegraphics[width=1\textwidth]{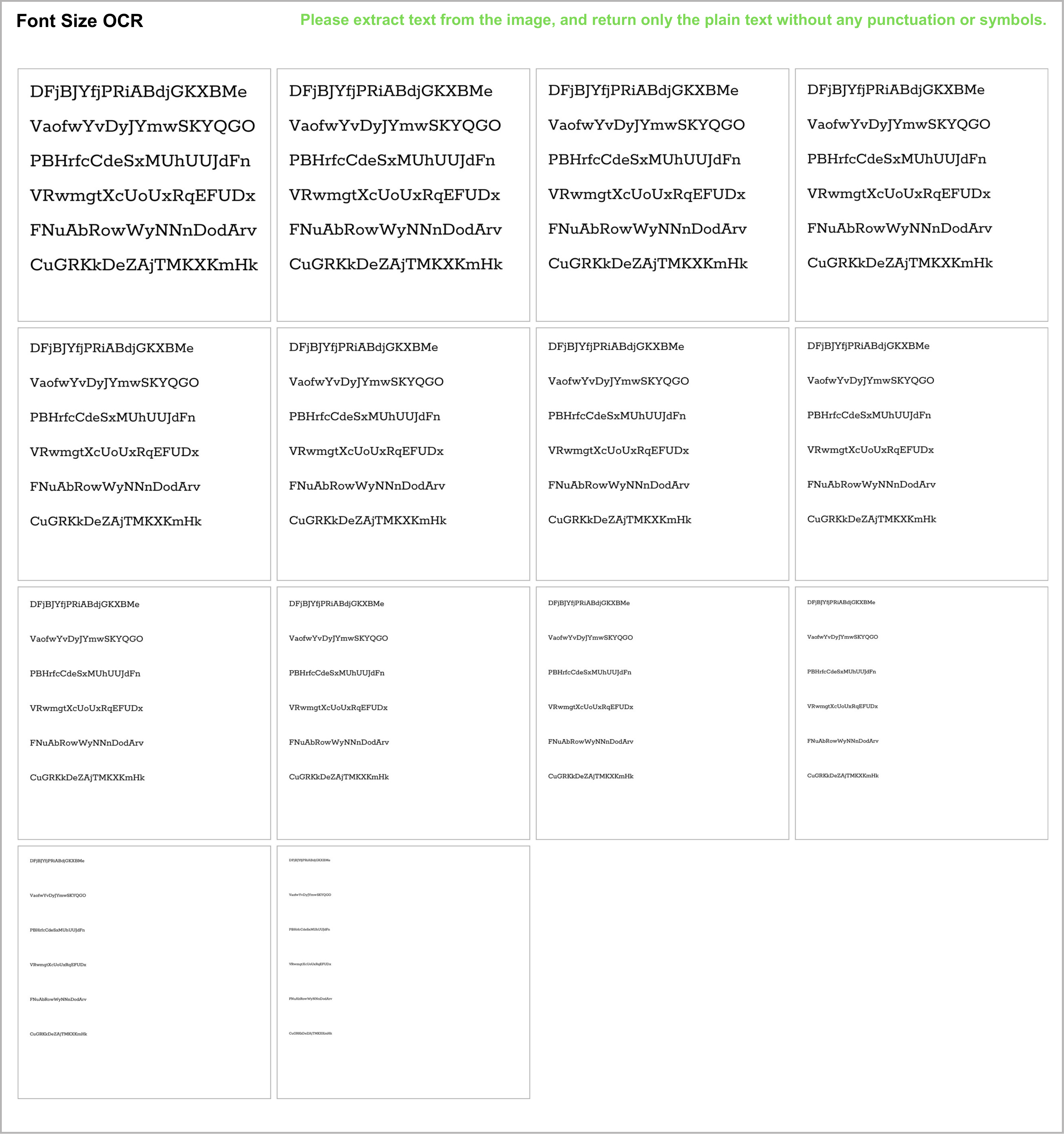}
\end{figure}
\vspace{-6mm}
\begin{figure}[H]
    \centering
    \includegraphics[width=1\textwidth]{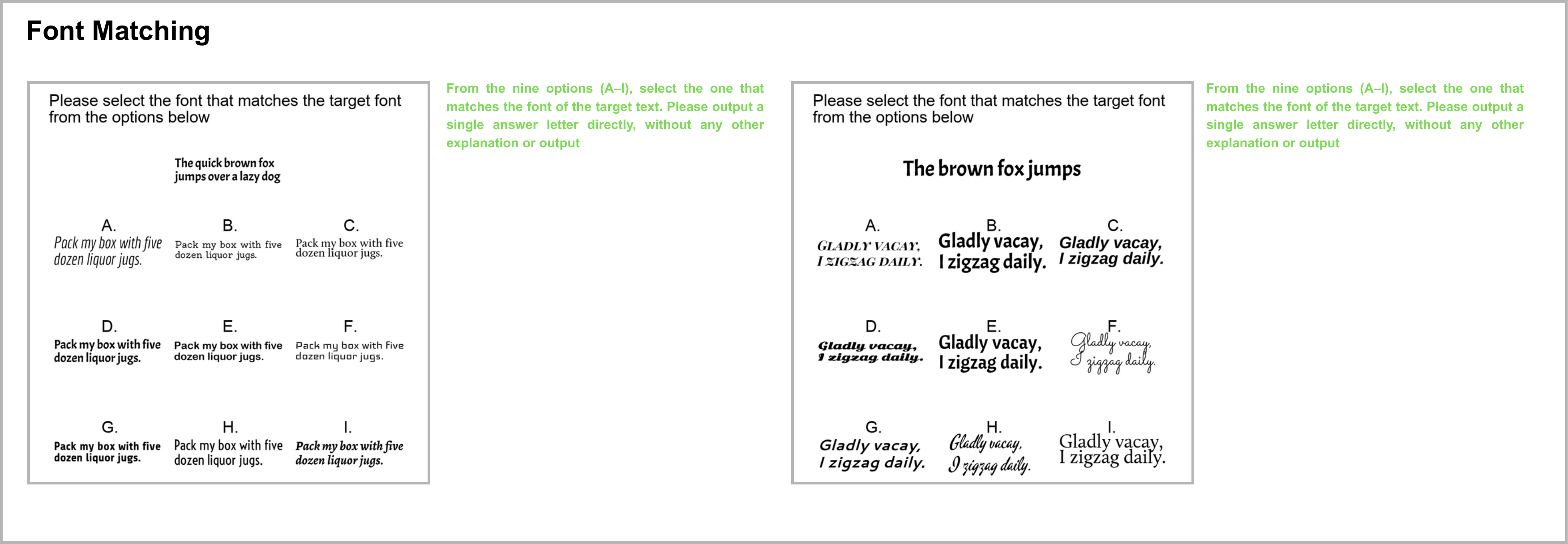}
\end{figure}
\vspace{-6mm}
\begin{figure}[H]
    \centering
    \includegraphics[width=1\textwidth]{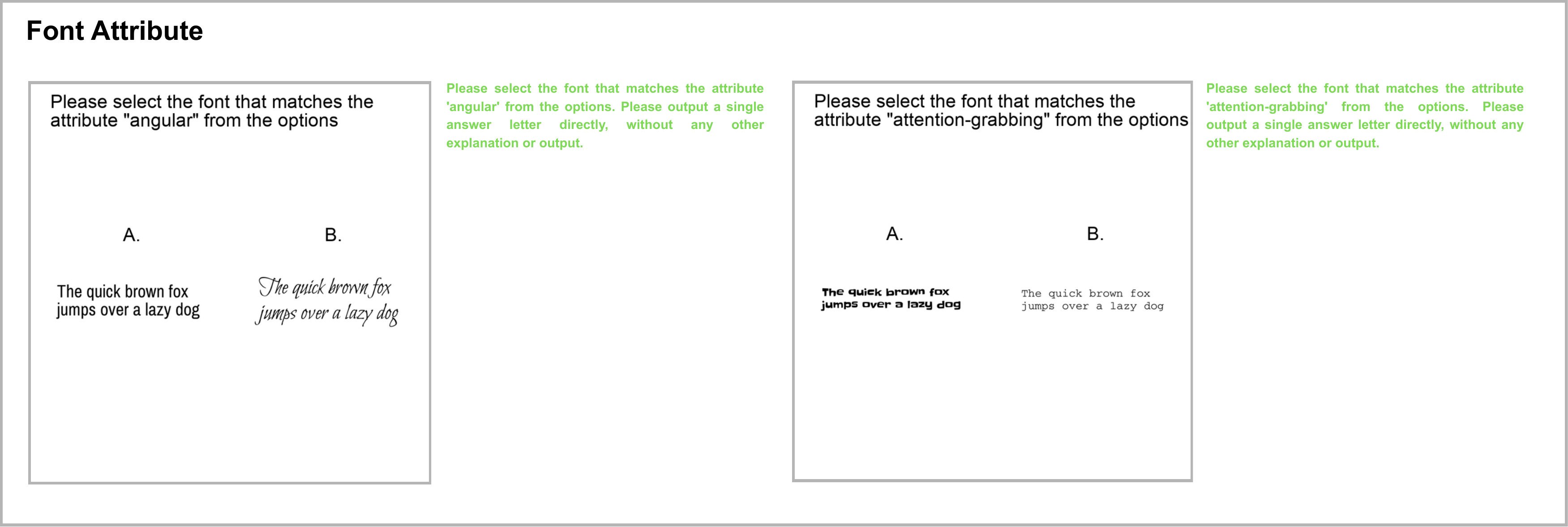}
\end{figure}
\vspace{-6mm}
\begin{figure}[H]
    \centering
    \includegraphics[width=1\textwidth]{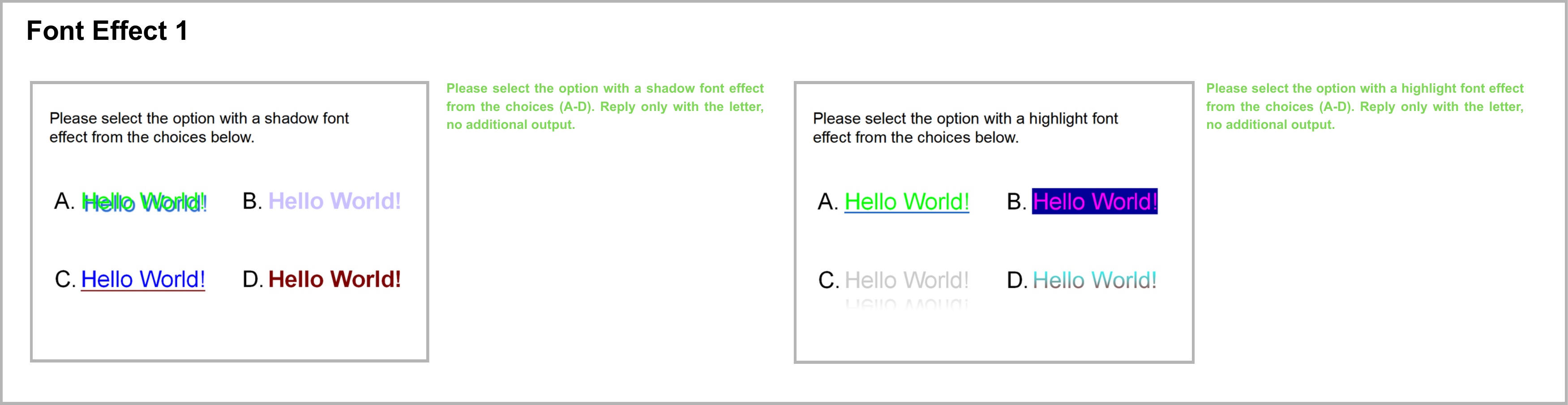}
\end{figure}
\vspace{-6mm}
\begin{figure}[H]
    \centering
    \includegraphics[width=1\textwidth]{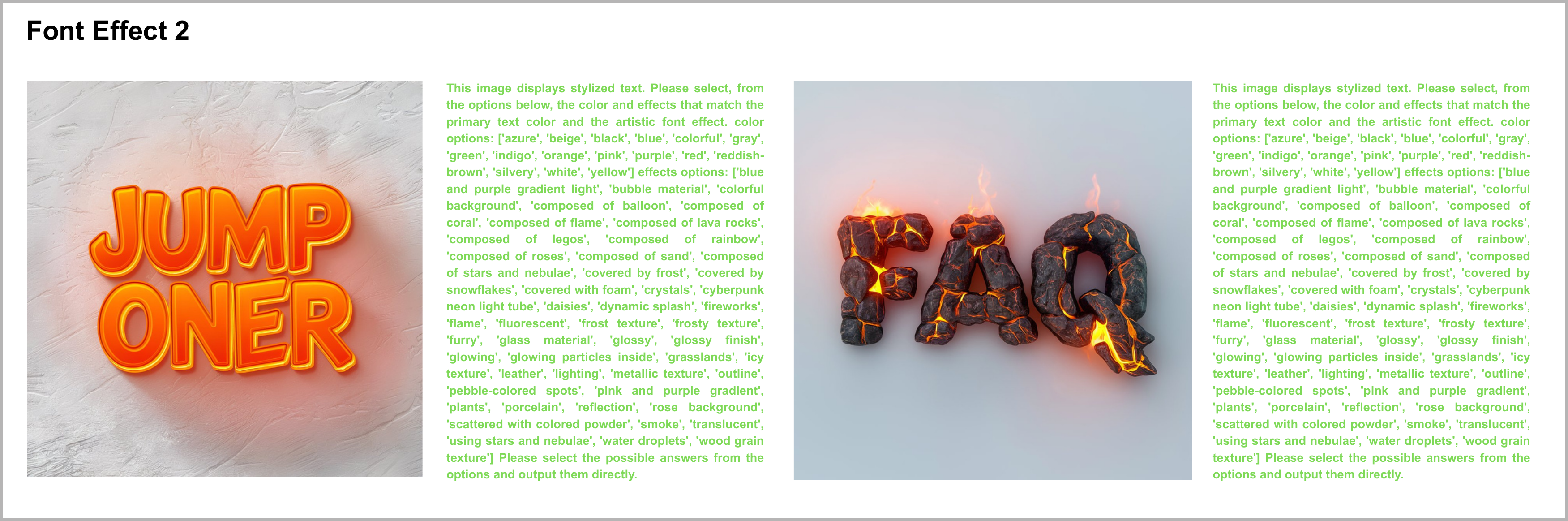}
\end{figure}
\vspace{-6mm}
\begin{figure}[H]
    \centering
    \includegraphics[width=1\textwidth]{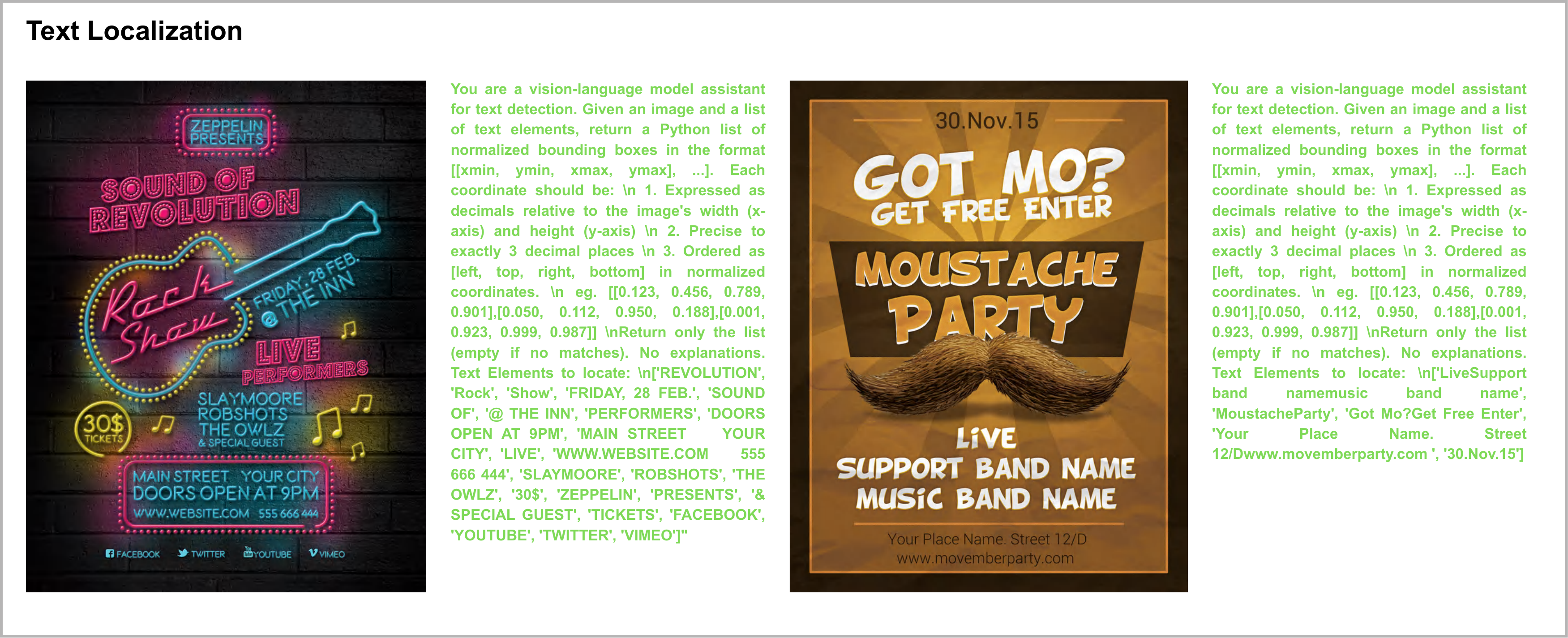}
\end{figure}
\vspace{-6mm}
\begin{figure}[H]
    \centering
    \includegraphics[width=1\textwidth]{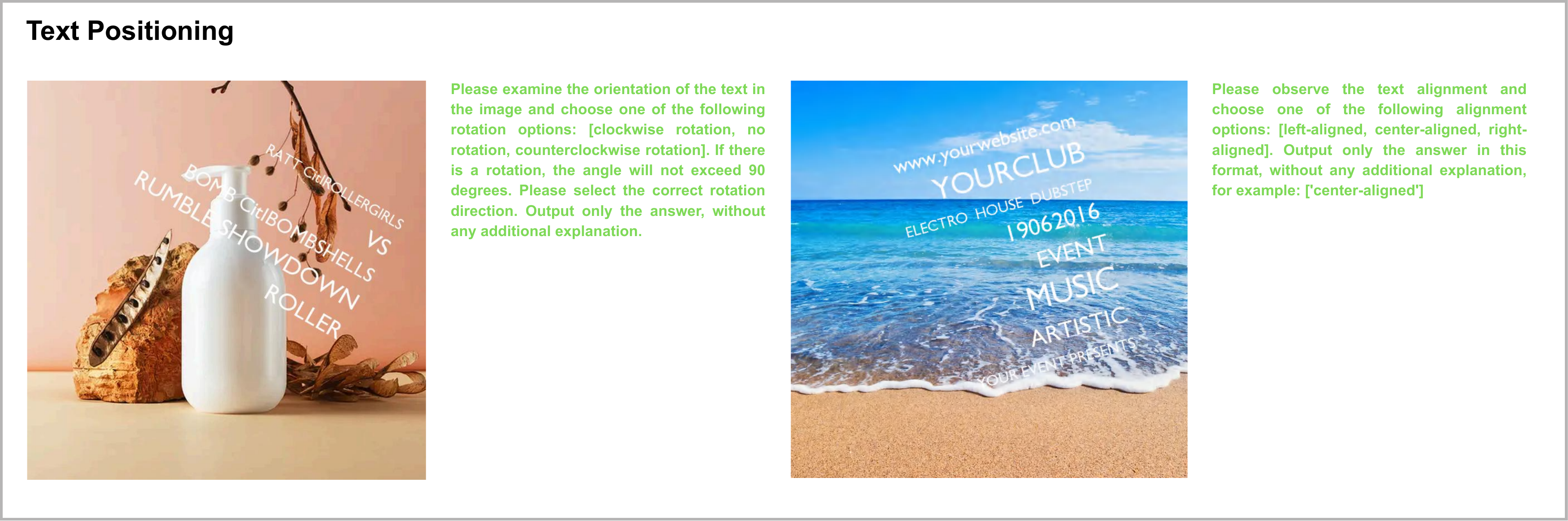}
\end{figure}
\vspace{-6mm}
\begin{figure}[H]
    \centering
    \includegraphics[width=1\textwidth]{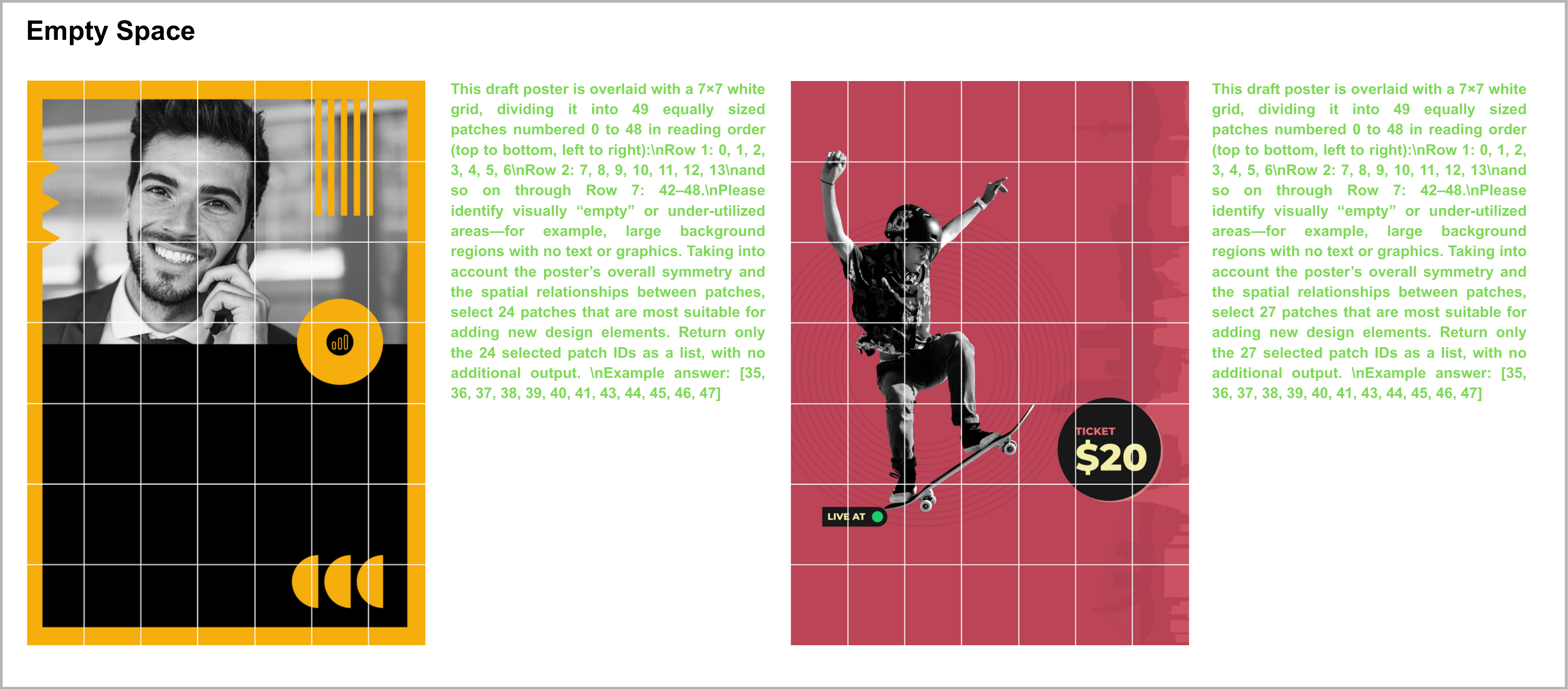}
\end{figure}
\vspace{-6mm}
\begin{figure}[H]
    \centering
    \includegraphics[width=1\textwidth]{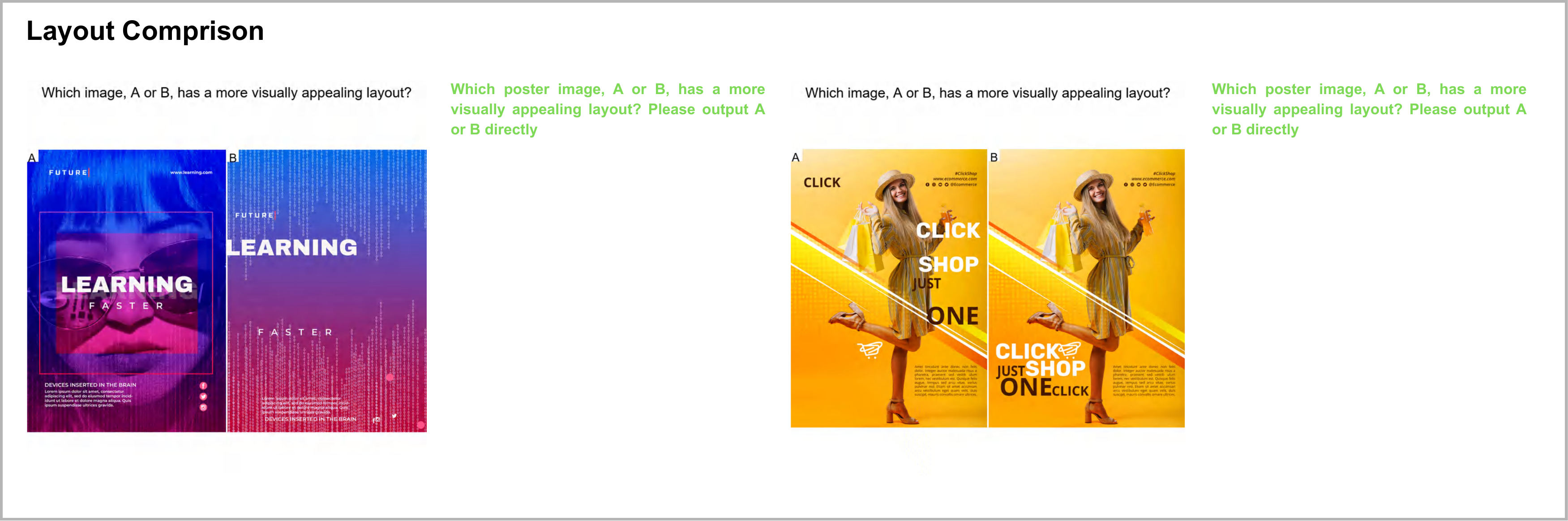}
\end{figure}
\vspace{-6mm}
\begin{figure}[H]
    \centering
    \includegraphics[width=1\textwidth]{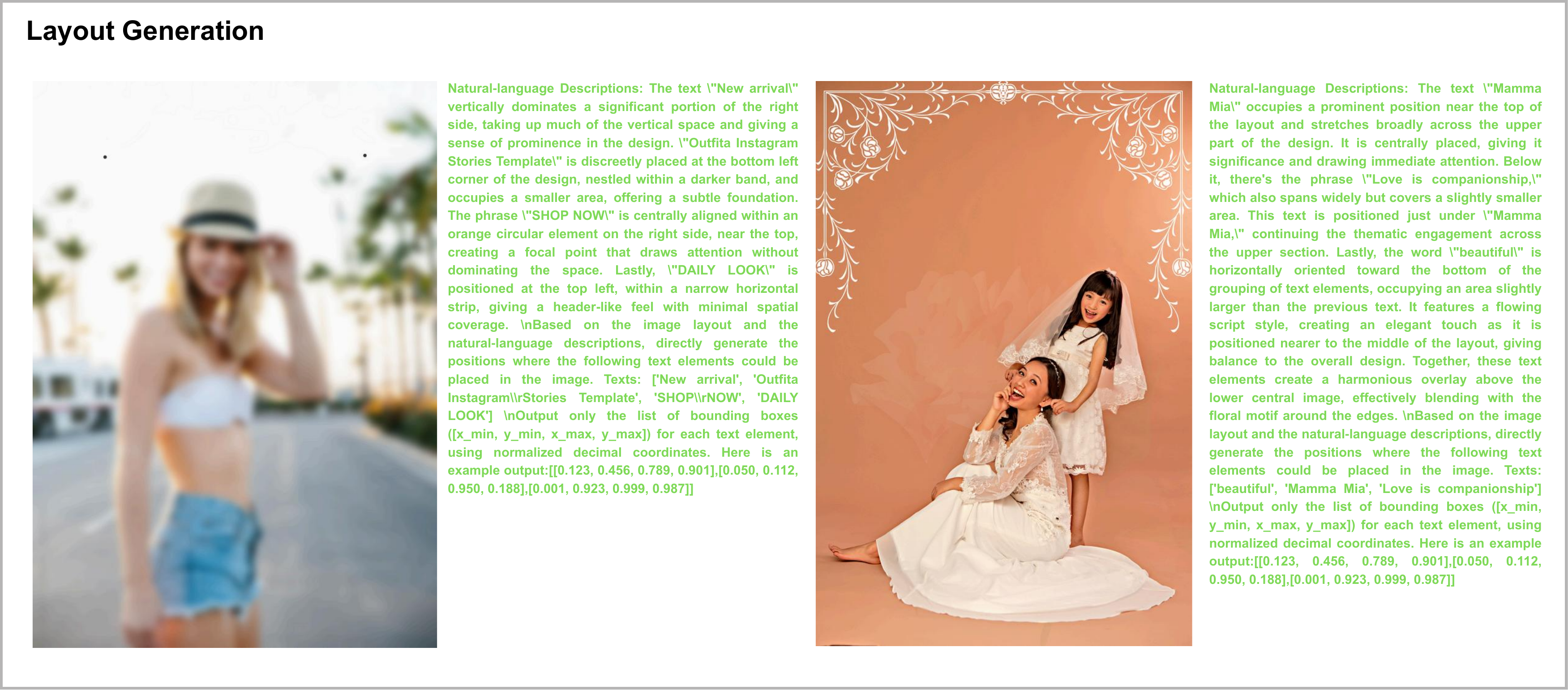}
\end{figure}
\vspace{-6mm}
\begin{figure}[H]
    \centering
    \includegraphics[width=1\textwidth]{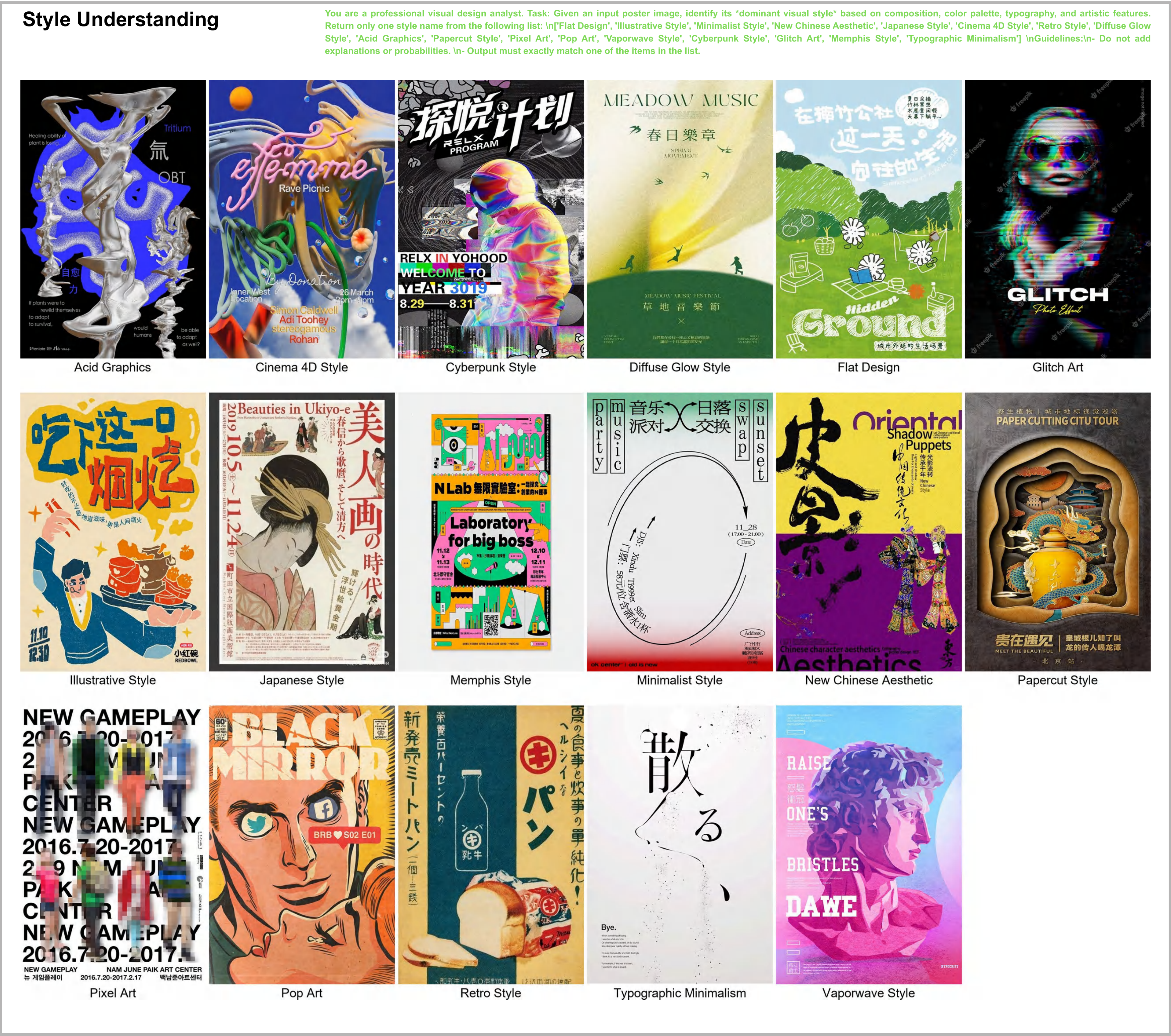}
\end{figure}
\vspace{-6mm}
\begin{figure}[H]
    \centering
    \includegraphics[width=1\textwidth]{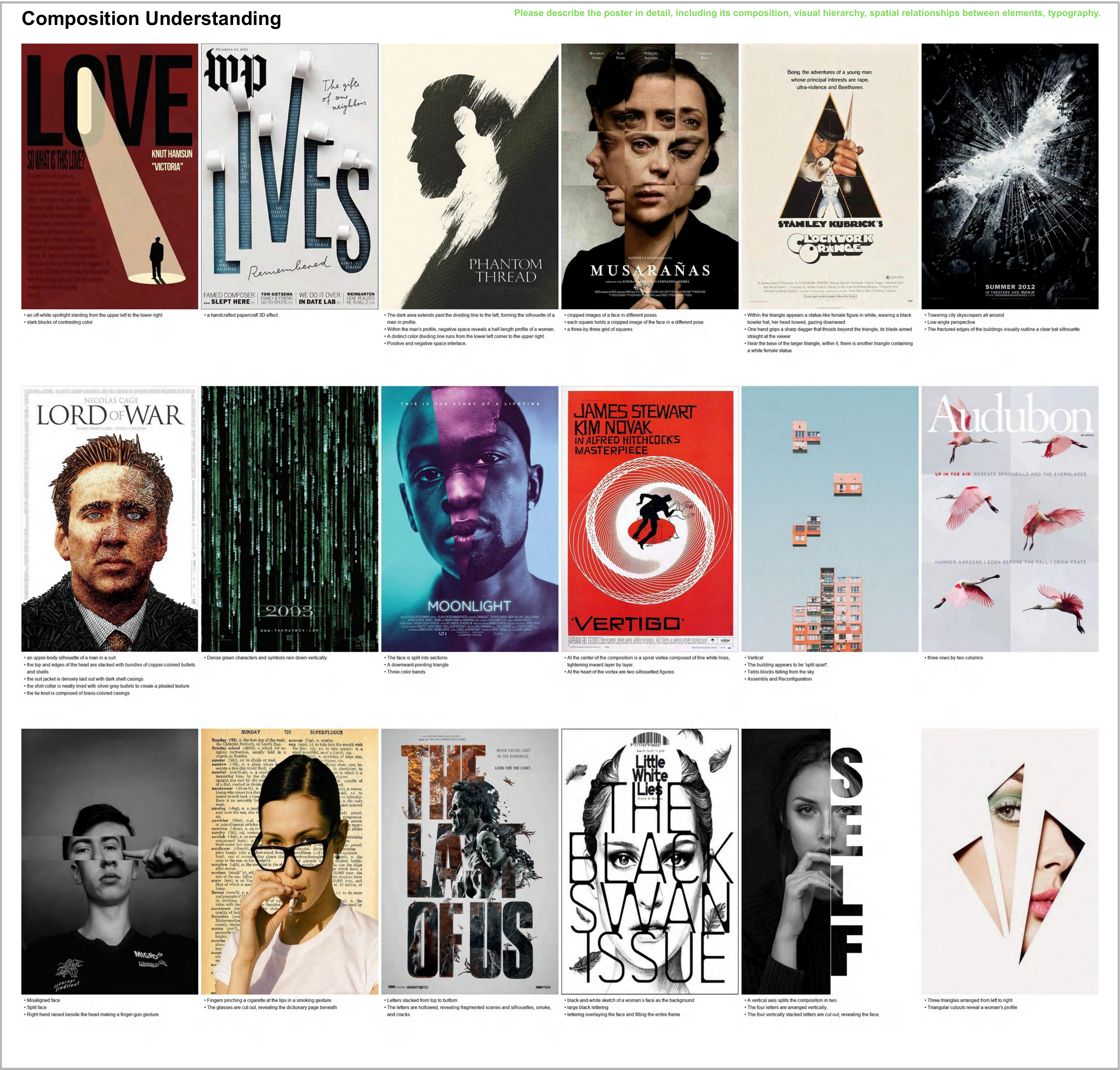}
\end{figure}
\vspace{-6mm}
\begin{figure}[H]
    \centering
    \includegraphics[width=1\textwidth]{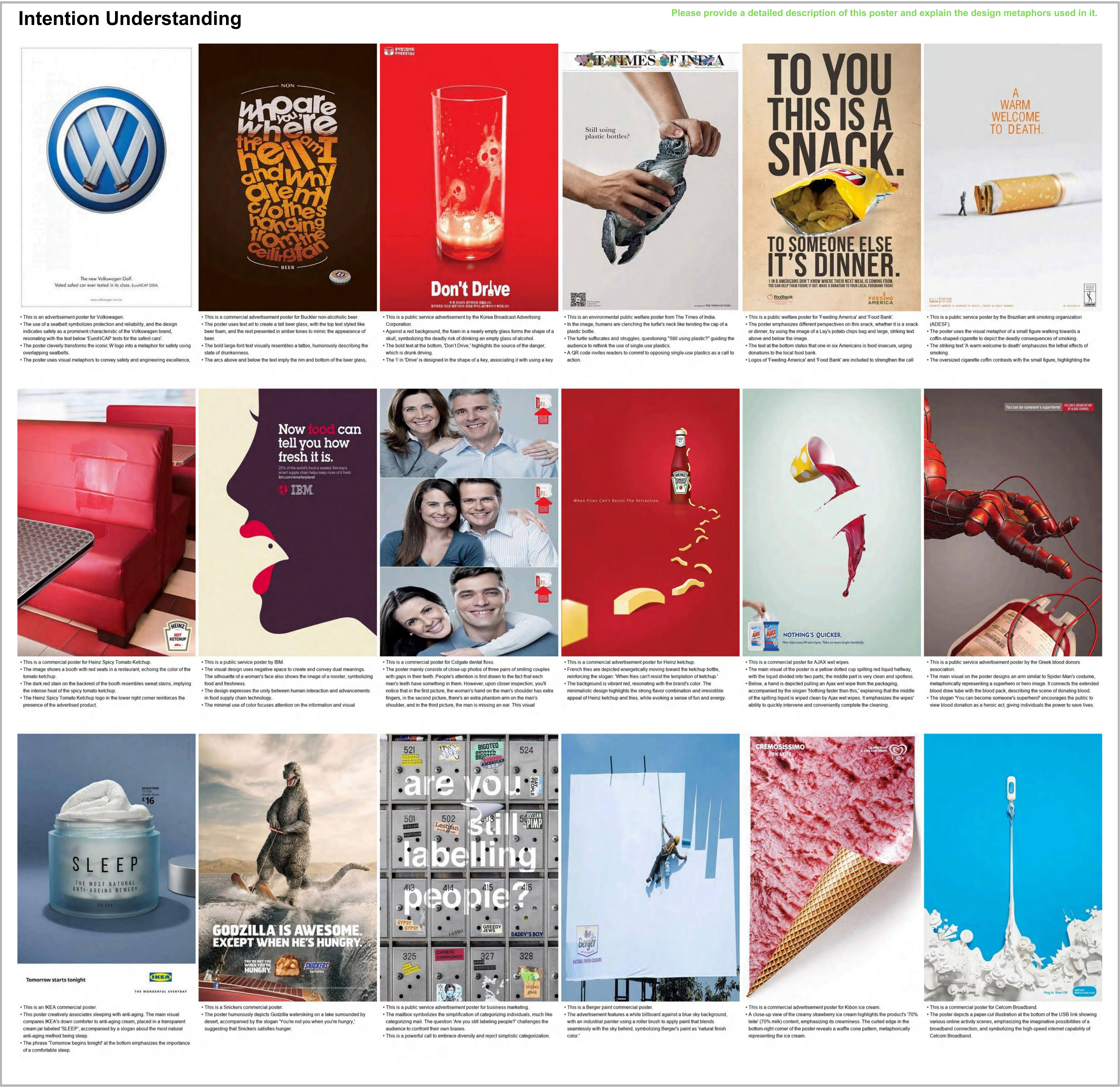}
\end{figure}
\vspace{-6mm}
\begin{figure}[H]
    \centering
    \includegraphics[width=1\textwidth]{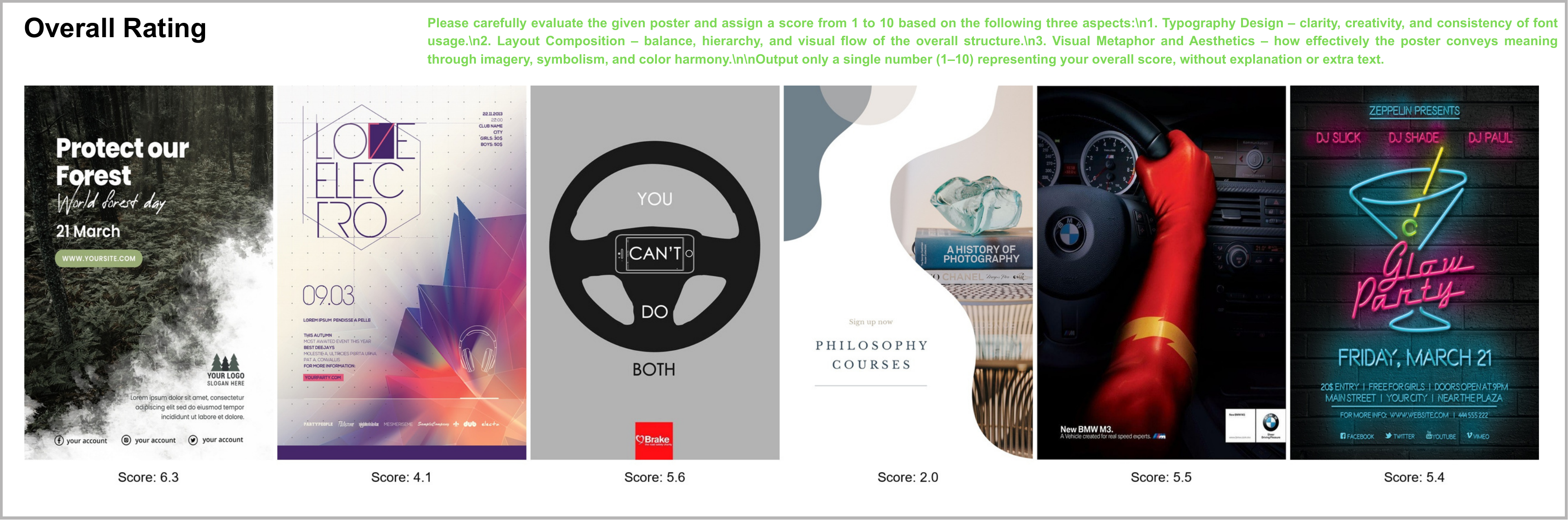}
\end{figure}
\vspace{-6mm}
\clearpage
\section{Generation Task Results}

\begin{figure}[H]
    \centering
    \includegraphics[width=1\textwidth]{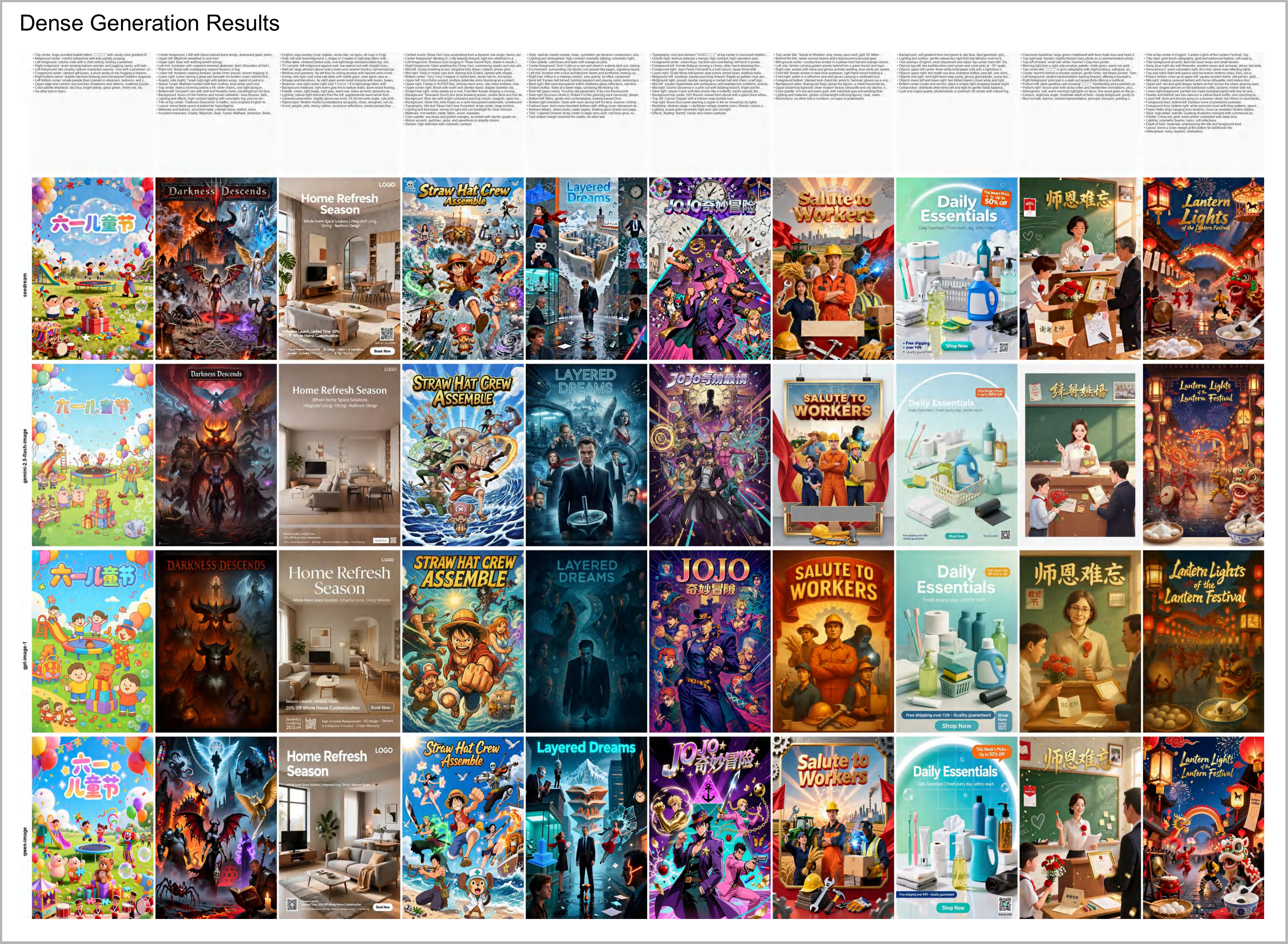}
\end{figure}

\begin{figure}[H]
    \centering
    \includegraphics[width=1\textwidth]{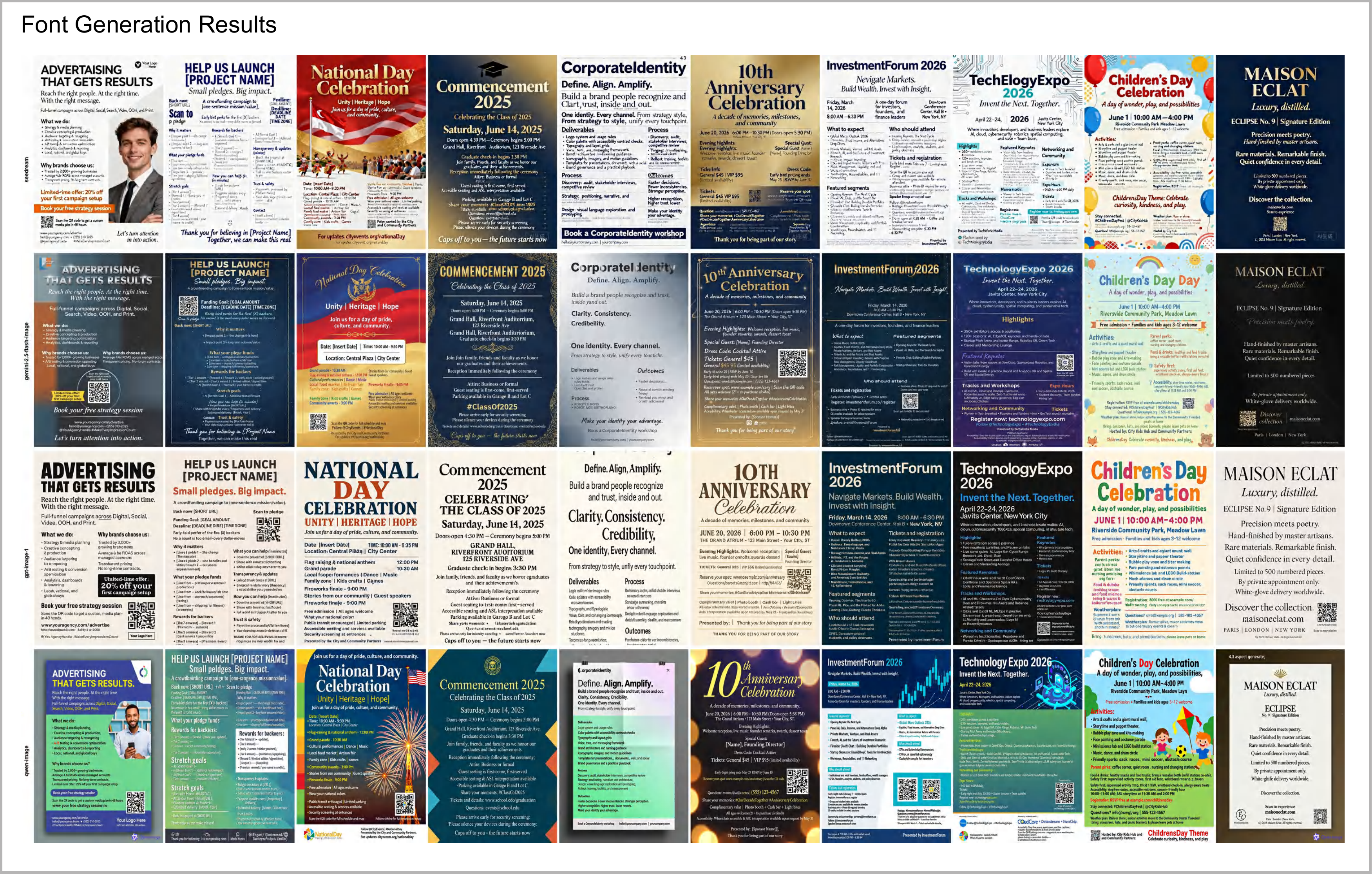}
\end{figure}
\begin{figure}[H]
    \centering
    \includegraphics[width=1\textwidth]{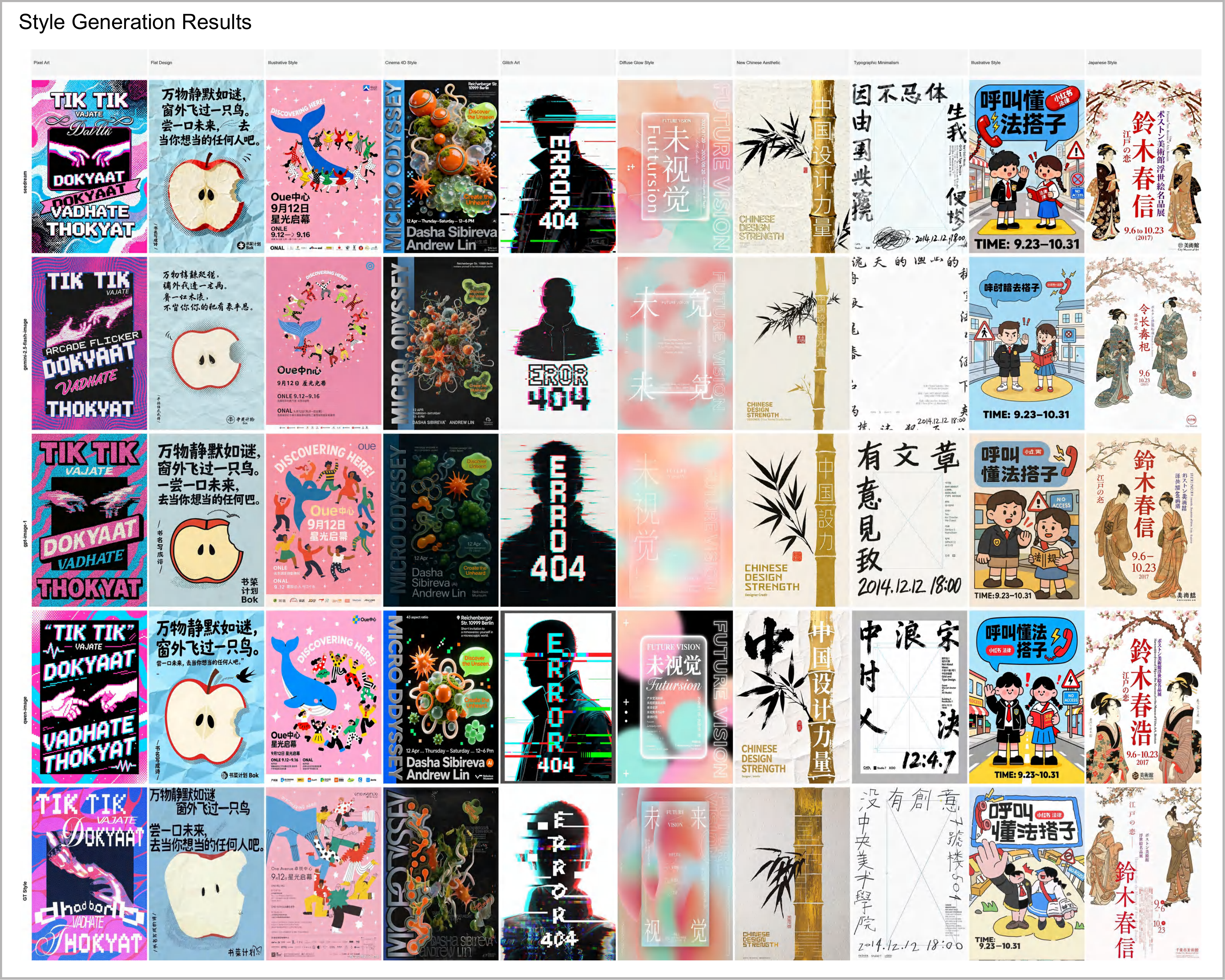}
\end{figure}
\begin{figure}[H]
    \centering
    \includegraphics[width=1\textwidth]{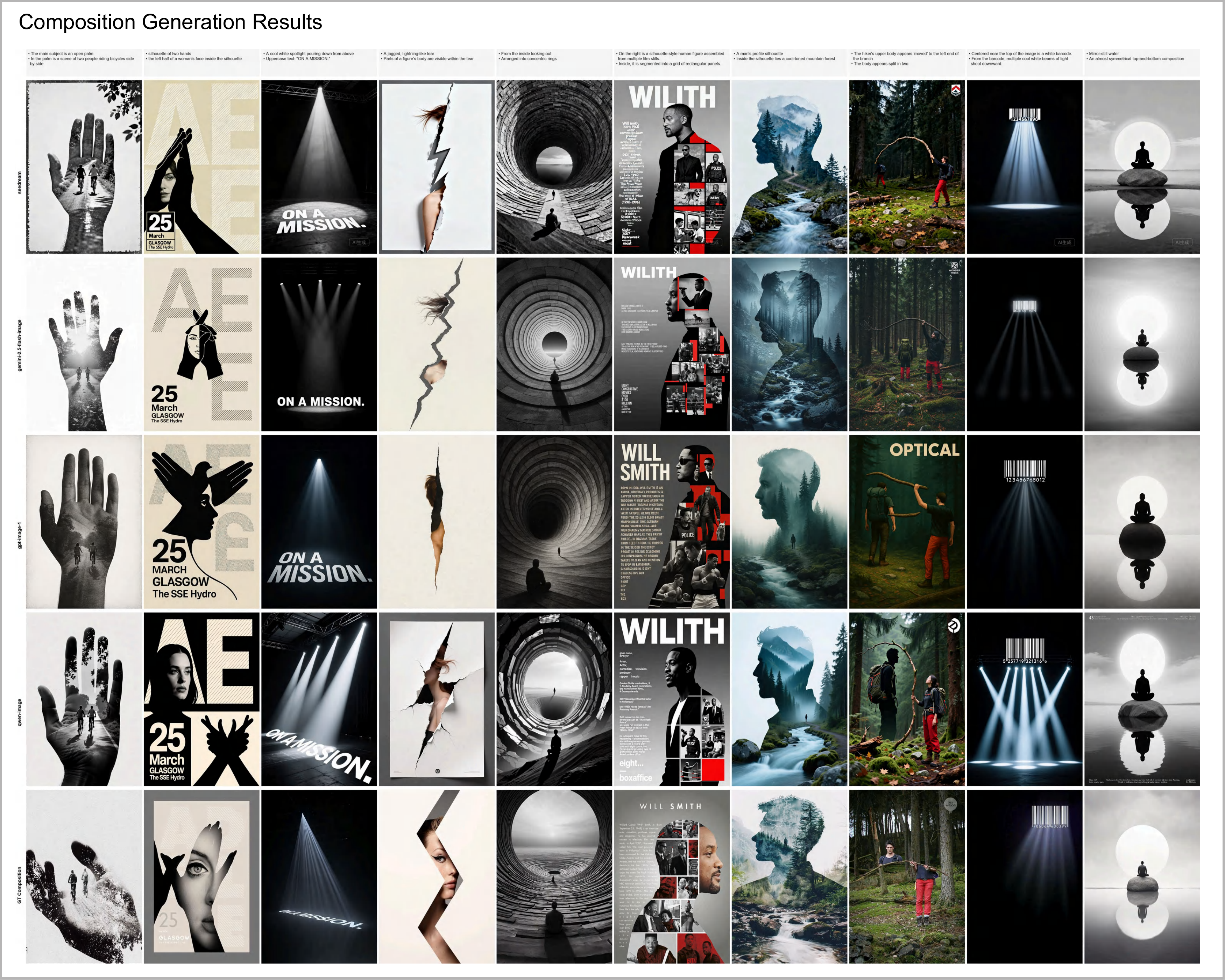}
\end{figure}
\begin{figure}[H]
    \centering
    \includegraphics[width=1\textwidth]{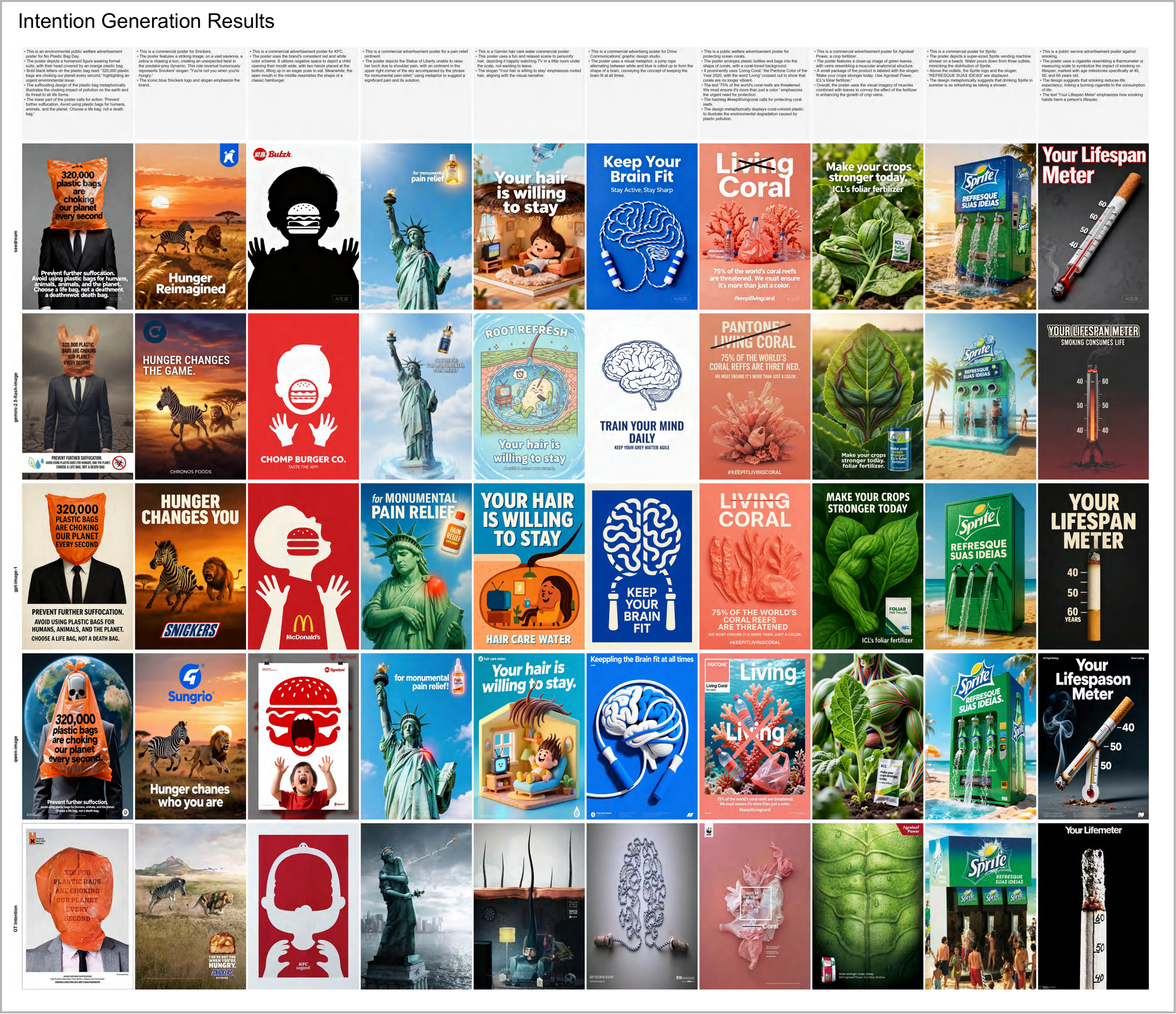}
\end{figure}

\clearpage

\clearpage

\end{document}